\documentclass[fleqn,10pt]{wlscirep}
\usepackage[utf8]{inputenc}
\usepackage[T1]{fontenc}
\usepackage{lineno}
\usepackage{multirow}
\usepackage{xurl}
\usepackage{xcolor}
\usepackage{bm}

\title{How accurate are existing land cover maps for agriculture in Sub-Saharan Africa?}

\author[1,*]{Hannah Kerner}
\author[2]{Catherine Nakalembe}
\author[3]{Adam Yang}
\author[2]{Ivan Zvonkov}
\author[2]{Ryan McWeeny}
\author[4]{Gabriel Tseng}
\author[2]{Inbal Becker-Reshef}
\affil[1]{Arizona State University, School of Computing and Augmented Intelligence, Tempe, AZ 85281, USA}
\affil[2]{University of Maryland, Department of Geographical Sciences, College Park, MD 20740, USA}
\affil[3]{University of Maryland, Department of Computer Science, College Park, MD 20740, USA}
\affil[4]{McGill University, Mila – Quebec AI Institute, Montreal, Quebec, Canada}

\affil[*]{corresponding author(s): Hannah Kerner (hkerner@asu.edu)}

\begin{abstract}
Satellite Earth observations (EO) can provide affordable and timely information for assessing crop conditions and food production. Such monitoring systems are essential in Africa, where food insecurity is high and agricultural statistics are sparse. EO-based monitoring systems require accurate cropland maps to provide information about croplands, but there is a lack of data to determine which of the many available land cover maps most accurately identify cropland in African countries. This study provides a quantitative evaluation and intercomparison of 11 publicly available land cover maps to assess their suitability for cropland classification and EO-based agriculture monitoring in Africa using statistically rigorous reference datasets from 8 countries. We hope the results of this study will help users determine the most suitable map for their needs and encourage future work to focus on resolving inconsistencies between maps and improving accuracy in low-accuracy regions. 
\end{abstract}
\begin{document}

\flushbottom
\maketitle

\thispagestyle{empty}

\section*{Introduction}

Africa is a critical area for research on food security. Half of the low-income and food-deficient countries (43 out of 86) are on the continent\cite{bjornlund2022food} and Africa has the highest prevalence of undernourished people (20.2\% of the population in 2021).\cite{wfp2022state} Efforts to assess, monitor, and mitigate food insecurity in Africa are hindered by a lack of information to inform such efforts, such as agricultural production statistics or crop conditions assessments. Satellite Earth observations (EO) provide an affordable, reliable, and timely source of information for assessing crop conditions and food production, but EO-based monitoring systems require accurate cropland masks that focus observations on locations where crops are likely being grown in a given year.\cite{nakalembe2023considerations,fritz2013need,fritz2019comparison} 

Cropland masks are typically derived from land cover maps, which can be agriculture-specific maps indicating cropland areas or more generic maps capturing multiple land cover and land use classes.
Map products are becoming the go-to source for informing agricultural policies, development investments, food and nutrition security monitoring, and climate modeling.\cite{mbow2017special} While EO-based map assessments are more scalable and affordable than traditional survey-based estimates, they must be used cautiously when they inform policy or other high-stakes decisions.\cite{kelly2018sustainability} Crop maps are directly used in EO-based monitoring systems such as the GEOGLAM Crop Monitors,\cite{becker2019geoglam,becker2020strengthening} Global Agricultural Monitoring (GLAM) System,\cite{becker2010monitoring} ASAP,\cite{rembold2019asap} and others.\cite{fritz2019comparison} Inaccurate crop masks increase the risk of over- or under-estimation of crop conditions or impacts of disasters or climate shocks on food production and may result in dubious statistics informing aid and development efforts.

 Production of global land cover maps continues to accelerate due to advancements in satellite sensors, machine learning, and cloud computing, as well as increased availability of open data and growing interest in environmental monitoring and management. Users now have an overwhelming array of options for land cover maps but lack independent quantitative information to evaluate these maps and assess their relative advantages or disadvantages, particularly in the context of agricultural monitoring in Africa. As more researchers and decision-makers look to Earth observations as a solution for agriculture and food security monitoring in Africa, there is a need for a comprehensive evaluation and inter-comparison of available maps to help users identify suitable data products and guide future work efforts to develop improved maps. 

 Previous work showed that available land cover maps had low accuracy in Africa and substantial inconsistencies between maps.\cite{nabil2020assessing,perez2017comparison,venter2022global,fritz2010comparison,fritz2008identifying,xu2019comparisons,vancutsem2012harmonizing,tchuente2011comparison,herold2008some} For example, Nabil et al. (2019) found that cropland accuracies for four maps were below 65\% in Africa.\cite{nabil2020assessing} Pérez-Hoyos et al. (2017) found full agreement between seven land cover maps in just 2.15\% of Africa.\cite{perez2017comparison} Venter et al. (2022) compared three global land cover maps released since 2021 (Google's Dynamic World,\cite{brown2022dynamic} ESA's WorldCover,\cite{zanaga2022esa} and Esri's Land Use Land Cover\cite{karra2021global}) and found substantial inaccuracies.\cite{venter2022global} The discrepancies between cropland classifications by different maps have been attributed to several factors including heterogeneous land cover, cloud cover affecting data availability in remote sensing model inputs, and the small and fragmented nature of crop fields in Africa.\cite{nabil2020assessing} Some work has sought to resolve these inconsistencies and create improved regional maps by combining the classifications from several individual maps.\cite{nabil2022constructing,fritz2011cropland,perez2020integrating} Researchers have hypothesized that higher-resolution satellite datasets (e.g., 10 m/px Sentinel-2) would enable the development of models that can capture small-scale and heterogeneous fields not captured by earlier models based on coarser datasets, but this hypothesis has not been tested as higher-resolution map products have been published. Most studies comparing land cover maps were published before several recent maps became available, focused on a small subset of maps, focused on broad land cover classes and not specifically agriculture, or used reference datasets sampled from large areas or lacking statistically rigorous sampling, and thus did not provide country-scale evaluations useful for assessing the suitability of maps for national crop monitoring.

 This study provides a quantitative evaluation and intercomparison of 11 publicly available land cover maps to assess their suitability for cropland classification and EO-based agriculture monitoring in Africa. We selected these maps to encompass a range of temporal availability (2009 to 2020), spatial resolutions (10 m/px to 1000 m/px), and classification approaches (tree-based to deep learning methods). We prioritized maps that were not included in previous studies comparing publicly available land cover maps. We summarize important metadata and cropland definitions for each map in Table \ref{tab:products-evaluated}. In order from finest to coarsest spatial resolution, the maps are Digital Earth Africa Cropland Extent,\cite{burton2022co} Dynamic World,\cite{brown2022dynamic} Esri LULC,\cite{karra2021global} ESA WorldCover,\cite{zanaga2022esa} CCI Land Cover Africa,\cite{cci_africa_feedback} GFSAD Global Cropland Extent,\cite{thenkabail2021global} Nabil et al.,\cite{nabil2022constructing} GLAD,\cite{potapov2022global} Copernicus Land Cover,\cite{buchhorn2020copernicus} ESA GlobCover,\cite{teamglobcover} and ASAP Crop Mask.\cite{rembold2019asap} The significant contributions of this study are as follows:
 \begin{itemize}
     \item We assessed the accuracy of 11 publicly available land cover maps using statistically rigorous reference datasets in eight Sub-Saharan African countries.
     \item We quantified and visualized cropland classification consensus across all maps as well as the pairwise agreement between individual maps.
     \item We assessed the correlation between map accuracy and spatial resolution as well as temporal relevance.
     \item We demonstrated how the choice of a map can affect downstream interpretations of agricultural conditions.
 \end{itemize}
 We evaluated the accuracy of each map using reference datasets for 8 countries selected to span the diverse agriculture of Sub-Saharan Africa (Kenya, Malawi, Mali, Rwanda, Tanzania, Togo, Uganda, and Zambia) and evaluation protocols based on best practices.\cite{stehman2019key} We quantified the consensus between maps using all pixels within each country boundary (after resampling all maps to a common resolution of 10 meters per pixel). In addition to our analysis, we make the following contributions:
 \begin{enumerate}
     \item A high quality reference dataset of 3,386 samples from 8 countries in Sub-Saharan Africa, which can be used as a common benchmark for assessing cropland accuracy in future land cover maps
     \item A Google Earth Engine App to enable users to visualize and compare the maps in this study
     \item A publicly-accessible code repository to facilitate evaluation for new countries or datasets not included in this study
 \end{enumerate}

We found very low consensus across the 11 compared maps when predicting cropland---all maps unanimously agree on a cropland prediction in fewer than 0.5\% of pixels in each of the 8 countries studied, which is much smaller than the estimated percentage of land used for agriculture in each country.\cite{faostat} There is a large disparity in the magnitude of the performance metrics between countries for all maps, with average F1 scores across all maps ranging from as low as $0.21 \pm 0.22$ for Mali to $0.71 \pm 0.16$ for Rwanda; the average F1 score is less than 0.7 for seven out of eight countries. We show that a majority vote ensemble that combines the predictions of all of the maps performs better than most, but not all, individual maps. We show that performance is overall weakly correlated with spatial resolution and temporal mismatch, especially for maps with resolution $\leq 100$ m/px or within 5 years of the reference data year. We hope this analysis will help users determine the most suitable map for their needs and encourage future work to focus on resolving inconsistencies between maps and improving accuracy in poorly classified regions.

\section*{Results}

\subsection*{Accuracy assessment}
\begin{figure}[ht]
\centering
\includegraphics[width=.9\linewidth]{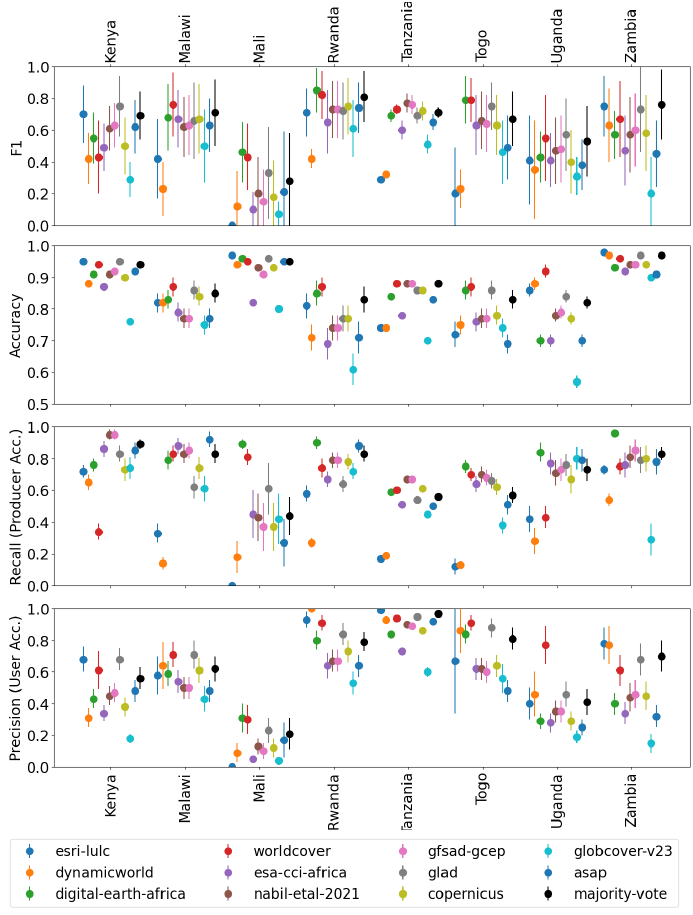}
\caption{Visual depiction of results in Table \ref{tab:metrics}. Some error bars are not visible due to small errors.}
\label{fig:results-visual}
\end{figure}

Table \ref{tab:metrics} reports the overall accuracy, F1 score, precision (user's accuracy), and recall (producer's accuracy) with associated standard errors from evaluating each of the maps using the reference datasets summarized in Table \ref{tab:evalsets}. Refer to the Methods section for definitions of each metric and derivation of standard errors. We also present the results from the majority vote ensemble of all maps. Figure \ref{fig:results-visual} shows a visual illustration of the results in Table \ref{tab:metrics}. 

Given results from four different metrics and 8 different countries, how can we conclude which map is ``best'' overall? The map that most frequently has the highest score across all metrics and countries is Digital Earth Africa (9 bold blue values in Table \ref{tab:metrics}, not including the Mean), followed by WorldCover (6 bold blue values). The map that most frequently has the highest or second-highest score across all metrics and countries is WorldCover (15 bold blue or bold black values in Table \ref{tab:metrics}, not including the Mean). However, summarizing the results in this way does not account for the disparities between metrics that can indicate poor overall performance. For example, Digital Earth Africa has the highest recall in Zambia and Uganda, but low precision scores in those countries. Esri has the highest accuracy in Mali, but all other metrics are 0 for the same dataset. Since only about 2\% of the points have a true label of crop in the Mali dataset, accuracy poorly describes the overall performance, evidenced by the Esri map achieving 0.98 accuracy even though it classified all of the reference points as non-crop. 

F1 score is often considered a better metric than overall accuracy to describe the overall performance because it combines two more descriptive metrics, precision and recall. The Majority Vote ensemble, WorldCover, and GLAD achieved the highest mean F1 score over all countries, closely followed by Digital Earth Africa. WorldCover achieved the highest mean precision (user's accuracy) score, followed by GLAD. Digital Earth Africa achieved the highest mean recall (producer's accuracy) score, followed by GFSAD and Nabil et al.\cite{nabil2022constructing} 

Considering the number of mean metrics for which a map scores highest or second highest, WorldCover comes in first (highest metric for 3 of 4 metrics) and GLAD comes in second (highest or second-highest metric for 3 of 4 metrics), Digital Earth Africa and the Majority Vote come in third (highest or second-highest metric for 2 out of 4 metrics). The lowest-performing maps were Copernicus, GlobCover, ASAP, ESA-CCI, and Dynamic World.

There is a large disparity in the magnitude of the performance metrics between countries. The last column of Table \ref{tab:metrics} reports each metric averaged across all maps for each country evaluated. The average F1 score ranges from as low as $0.21 \pm 0.22$ for Mali to $0.71 \pm 0.16$ for Rwanda. The average F1 score is below 0.7 for all countries except Rwanda. The precision score is also very low for some countries, particularly Mali, with an average precision of $0.15 \pm 0.06$. This illustrates that even though some maps may perform better than others, accurate cropland classification in these countries remains challenging.

\begin{table}[h]
\resizebox{\textwidth}{!}{\begin{tabular}{@{}cccccccccccccc|c@{}}
\toprule
Country & Metric & Majority Vote & Esri & Dynamic World & DEA & WorldCover & ESA-CCI & Nabil et al.\cite{nabil2022constructing} & GFSAD & GLAD & Copernicus & GlobCover & ASAP & Mean \\ \toprule

\multirow{4}{*}{Kenya} & Accuracy & $\bm{0.94\pm0.01}$ & \color{blue}$\bm{0.95\pm0.01}$ & $0.88\pm0.01$ & $0.91\pm0.01$ & $\bm{0.94\pm0.01}$ & $0.87\pm0.01$ & $0.91\pm0.01$ & $0.92\pm0.01$ & \color{blue}$\bm{0.95\pm0.01}$ & $0.90\pm0.01$ & $0.76\pm0.01$ & $0.92\pm0.01$ & $0.90\pm0.01$ \\ 
& F1 & $0.69\pm0.15$ & $\bm{0.70\pm0.18}$ & $0.42\pm0.16$ & $0.55\pm0.16$ & $0.43\pm0.23$ & $0.49\pm0.15$ & $0.61\pm0.14$ & $0.63\pm0.14$ & \color{blue}$\bm{0.75\pm0.19}$ & $0.50\pm0.18$ & $0.29\pm0.11$ & $0.62\pm0.17$ & $0.56\pm0.16$ \\ 
& Precision (UA) & $0.56\pm0.07$ & $\bm{0.68\pm0.08}$ & $0.31\pm0.06$ & $0.43\pm0.06$ & $0.61\pm0.12$ & $0.34\pm0.05$ & $0.45\pm0.06$ & $0.47\pm0.06$ & $\bm{0.68\pm0.07}$ & $0.38\pm0.06$ & $0.18\pm0.03$ & $0.48\pm0.07$ & $0.46\pm0.07$ \\ 
& Recall (PA) & $\bm{0.89\pm0.03}$ & $0.72\pm0.04$ & $0.65\pm0.05$ & $0.76\pm0.04$ & $0.34\pm0.05$ & $0.86\pm0.05$ & \color{blue}$\bm{0.95\pm0.03}$ & \color{blue}$\bm{0.95\pm0.03}$ & $0.83\pm0.05$ & $0.73\pm0.07$ & $0.74\pm0.07$ & $0.85\pm0.05$ & $0.77\pm0.05$ \\ 
\midrule 
\multirow{4}{*}{Malawi} & Accuracy & $0.85\pm0.03$ & $0.82\pm0.03$ & $0.82\pm0.03$ & $0.83\pm0.03$ & $\bm{0.87\pm0.03}$ & $0.79\pm0.03$ & $0.77\pm0.03$ & $0.77\pm0.03$ & $0.86\pm0.03$ & $0.84\pm0.03$ & $0.75\pm0.03$ & $0.77\pm0.03$ & $0.81\pm0.03$ \\ 
& F1 & $0.71\pm0.21$ & $0.42\pm0.25$ & $0.23\pm0.17$ & $0.68\pm0.21$ & $0.76\pm0.20$ & $0.67\pm0.18$ & $0.62\pm0.19$ & $0.63\pm0.19$ & $0.66\pm0.24$ & $0.67\pm0.22$ & $0.50\pm0.23$ & $0.63\pm0.17$ & $0.60\pm0.20$ \\ 
& Precision (UA) & $0.62\pm0.08$ & $0.58\pm0.12$ & $0.64\pm0.15$ & $0.59\pm0.08$ & $\bm{0.71\pm0.08}$ & $0.54\pm0.07$ & $0.50\pm0.07$ & $0.50\pm0.07$ & $\bm{0.71\pm0.09}$ & $0.61\pm0.08$ & $0.43\pm0.08$ & $0.48\pm0.07$ & $0.58\pm0.09$ \\ 
& Recall (PA) & $0.83\pm0.06$ & $0.33\pm0.06$ & $0.14\pm0.04$ & $0.79\pm0.06$ & $0.83\pm0.05$ & $0.88\pm0.05$ & $0.83\pm0.06$ & $0.85\pm0.05$ & $0.62\pm0.07$ & $0.74\pm0.07$ & $0.61\pm0.08$ & $\bm{0.92\pm0.05}$ & $0.70\pm0.06$ \\ 
\midrule 
\multirow{4}{*}{Mali} & Accuracy & $0.95\pm0.01$ & \color{blue}$\bm{0.97\pm0.01}$ & $0.94\pm0.01$ & $\bm{0.96\pm0.01}$ & $0.95\pm0.01$ & $0.82\pm0.01$ & $0.93\pm0.01$ & $0.91\pm0.01$ & $\bm{0.96\pm0.01}$ & $0.93\pm0.01$ & $0.80\pm0.01$ & $0.95\pm0.01$ & $0.92\pm0.01$ \\ 
& F1 & $0.28\pm0.30$ & $0.00\pm 0.00$ & $0.12\pm0.22$ & \color{blue}$\bm{0.46\pm0.19}$ & $0.43\pm0.21$ & $0.10\pm0.11$ & $0.20\pm0.23$ & $0.15\pm0.20$ & $0.33\pm0.29$ & $0.18\pm0.23$ & $0.07\pm0.08$ & $0.21\pm0.38$ & $0.21\pm0.22$ \\ 
& Precision (UA) & $0.21\pm0.10$ & $0.00\pm0.00$ & $0.09\pm0.06$ & \color{blue}$\bm{0.31\pm0.09}$ & $\bm{0.30\pm0.09}$ & $0.05\pm0.02$ & $0.13\pm0.05$ & $0.10\pm0.05$ & $0.23\pm0.08$ & $0.12\pm0.06$ & $0.04\pm0.02$ & $0.17\pm0.11$ & $0.15\pm0.06$ \\ 
& Recall (PA) & $0.44\pm0.12$ & $0.00\pm0.00$ & $0.18\pm0.10$ & \color{blue}$\bm{0.89\pm0.03}$ & $\bm{0.81\pm0.05}$ & $0.45\pm0.15$ & $0.43\pm0.15$ & $0.37\pm0.15$ & $0.61\pm0.16$ & $0.37\pm0.15$ & $0.42\pm0.16$ & $0.27\pm0.15$ & $0.44\pm0.11$ \\ 
\midrule 
\multirow{4}{*}{Rwanda} & Accuracy & $0.83\pm0.04$ & $0.81\pm0.04$ & $0.71\pm0.04$ & $\bm{0.85\pm0.04}$ & \color{blue}$\bm{0.87\pm0.03}$ & $0.69\pm0.05$ & $0.74\pm0.04$ & $0.74\pm0.04$ & $0.77\pm0.04$ & $0.77\pm0.04$ & $0.61\pm0.05$ & $0.71\pm0.05$ & $0.76\pm0.04$ \\ 
& F1 & $0.81\pm0.16$ & $0.71\pm0.15$ & $0.42\pm0.06$ & \color{blue}$\bm{0.85\pm0.14}$ & $\bm{0.82\pm0.15}$ & $0.65\pm0.20$ & $0.73\pm0.18$ & $0.73\pm0.18$ & $0.72\pm0.18$ & $0.75\pm0.18$ & $0.61\pm0.18$ & $0.74\pm0.16$ & $0.71\pm0.16$ \\ 
& Precision (UA) & $0.79\pm0.06$ & $\bm{0.93\pm0.05}$ & \color{blue}$\bm{1.00\pm0.00}$ & $0.80\pm0.06$ & $0.91\pm0.05$ & $0.64\pm0.08$ & $0.67\pm0.07$ & $0.67\pm0.07$ & $0.84\pm0.07$ & $0.73\pm0.07$ & $0.53\pm0.07$ & $0.64\pm0.07$ & $0.76\pm0.06$ \\ 
& Recall (PA) & $0.83\pm0.05$ & $0.58\pm0.05$ & $0.27\pm0.03$ & \color{blue}$\bm{0.90\pm0.04}$ & $0.74\pm0.06$ & $0.67\pm0.05$ & $0.79\pm0.05$ & $0.79\pm0.05$ & $0.64\pm0.05$ & $0.78\pm0.05$ & $0.72\pm0.05$ & $\bm{0.88\pm0.04}$ & $0.72\pm0.05$ \\ 
\midrule 
\multirow{4}{*}{Tanzania} & Accuracy & $\bm{0.88\pm0.01}$ & $0.74\pm0.01$ & $0.74\pm0.01$ & $0.84\pm0.01$ & $\bm{0.88\pm0.01}$ & $0.78\pm0.01$ & $\bm{0.88\pm0.01}$ & $\bm{0.88\pm0.01}$ & $0.86\pm0.01$ & $0.86\pm0.01$ & $0.70\pm0.01$ & $0.83\pm0.01$ & $0.82\pm0.01$ \\ 
& F1 & $0.71\pm0.03$ & $0.29\pm0.01$ & $0.32\pm0.02$ & $0.69\pm0.04$ & $0.73\pm0.03$ & $0.60\pm0.06$ & \color{blue}$\bm{0.77\pm0.06}$ & $\bm{0.76\pm0.06}$ & $0.69\pm0.05$ & $0.72\pm0.06$ & $0.51\pm0.06$ & $0.65\pm0.05$ & $0.62\pm0.04$ \\ 
& Precision (UA) & $\bm{0.97\pm0.01}$ & \color{blue}$\bm{0.99\pm0.01}$ & $0.93\pm0.02$ & $0.84\pm0.02$ & $0.94\pm0.01$ & $0.73\pm0.02$ & $0.90\pm0.02$ & $0.89\pm0.02$ & $0.95\pm0.01$ & $0.86\pm0.02$ & $0.60\pm0.03$ & $0.92\pm0.02$ & $0.88\pm0.02$ \\ 
& Recall (PA) & $0.56\pm0.01$ & $0.17\pm0.00$ & $0.19\pm0.00$ & $0.59\pm0.01$ & $0.60\pm0.01$ & $0.51\pm0.02$ & $\bm{0.67\pm0.02}$ & $\bm{0.67\pm0.02}$ & $0.54\pm0.02$ & $0.61\pm0.02$ & $0.45\pm0.02$ & $0.50\pm0.02$ & $0.51\pm0.01$ \\ 
\midrule 
\multirow{4}{*}{Togo} & Accuracy & $0.83\pm0.03$ & $0.72\pm0.04$ & $0.75\pm0.03$ & $\bm{0.86\pm0.03}$ & \color{blue}$\bm{0.87\pm0.03}$ & $0.76\pm0.03$ & $0.77\pm0.03$ & $0.77\pm0.03$ & $\bm{0.86\pm0.03}$ & $0.78\pm0.03$ & $0.74\pm0.03$ & $0.69\pm0.03$ & $0.78\pm0.03$ \\ 
& F1 & $0.67\pm0.17$ & $0.20\pm0.29$ & $0.23\pm0.12$ & \color{blue}$\bm{0.79\pm0.15}$ & \color{blue}$\bm{0.79\pm0.14}$ & $0.63\pm0.19$ & $0.66\pm0.18$ & $0.64\pm0.18$ & $0.75\pm0.15$ & $0.63\pm0.19$ & $0.46\pm0.20$ & $0.49\pm0.20$ & $0.58\pm0.18$ \\ 
& Precision (UA) & $0.81\pm0.07$ & $0.67\pm0.33$ & $0.86\pm0.14$ & $0.84\pm0.06$ & \color{blue}$\bm{0.91\pm0.05}$ & $0.62\pm0.07$ & $0.62\pm0.07$ & $0.60\pm0.07$ & $0.88\pm0.06$ & $0.64\pm0.07$ & $0.56\pm0.09$ & $0.48\pm0.07$ & $0.71\pm0.10$ \\ 
& Recall (PA) & $0.57\pm0.05$ & $0.12\pm0.05$ & $0.13\pm0.02$ & \color{blue}$\bm{0.75\pm0.04}$ & $\bm{0.70\pm0.04}$ & $0.64\pm0.05$ & $\bm{0.70\pm0.05}$ & $0.68\pm0.05$ & $0.66\pm0.05$ & $0.62\pm0.05$ & $0.38\pm0.05$ & $0.51\pm0.06$ & $0.54\pm0.05$ \\ 
\midrule 
\multirow{4}{*}{Uganda} & Accuracy & $0.82\pm0.02$ & $0.86\pm0.02$ & $0.88\pm0.02$ & $0.70\pm0.02$ & \color{blue}$\bm{0.92\pm0.02}$ & $0.70\pm0.02$ & $0.78\pm0.02$ & $0.79\pm0.02$ & $0.84\pm0.02$ & $0.77\pm0.02$ & $0.57\pm0.02$ & $0.70\pm0.02$ & $0.78\pm0.02$ \\ 
& F1 & $0.53\pm0.22$ & $0.41\pm0.28$ & $0.35\pm0.31$ & $0.43\pm0.16$ & $\bm{0.55\pm0.27}$ & $0.41\pm0.17$ & $0.47\pm0.21$ & $0.48\pm0.21$ & \color{blue}$\bm{0.57\pm0.23}$ & $0.40\pm0.20$ & $0.31\pm0.12$ & $0.38\pm0.16$ & $0.44\pm0.21$ \\ 
& Precision (UA) & $0.41\pm0.08$ & $0.40\pm0.10$ & $0.46\pm0.14$ & $0.29\pm0.05$ & \color{blue}$\bm{0.77\pm0.12}$ & $0.28\pm0.06$ & $0.35\pm0.07$ & $0.35\pm0.07$ & $0.46\pm0.08$ & $0.29\pm0.06$ & $0.19\pm0.04$ & $0.25\pm0.05$ & $0.38\pm0.08$ \\ 
& Recall (PA) & $0.73\pm0.07$ & $0.42\pm0.08$ & $0.28\pm0.08$ & \color{blue}$\bm{0.84\pm0.06}$ & $0.43\pm0.07$ & $0.77\pm0.07$ & $0.71\pm0.08$ & $0.73\pm0.07$ & $0.76\pm0.07$ & $0.67\pm0.09$ & $\bm{0.80\pm0.07}$ & $0.79\pm0.07$ & $0.66\pm0.07$ \\ 
\midrule 
\multirow{4}{*}{Zambia} & Accuracy & $\bm{0.97\pm0.01}$ & \color{blue}$\bm{0.98\pm0.01}$ & $\bm{0.97\pm0.01}$ & $0.93\pm0.01$ & $0.96\pm0.01$ & $0.92\pm0.01$ & $0.94\pm0.01$ & $0.94\pm0.01$ & $\bm{0.97\pm0.01}$ & $0.94\pm0.01$ & $0.90\pm0.01$ & $0.91\pm0.01$ & $0.94\pm0.01$ \\ 
& F1 & \color{blue}$\bm{0.76\pm0.22}$ & $\bm{0.75\pm0.19}$ & $0.63\pm0.23$ & $0.57\pm0.15$ & $0.67\pm0.24$ & $0.47\pm0.22$ & $0.57\pm0.24$ & $0.60\pm0.23$ & $0.73\pm0.27$ & $0.58\pm0.24$ & $0.20\pm0.23$ & $0.45\pm0.21$ & $0.58\pm0.22$ \\ 
& Precision (UA) & $0.70\pm0.10$ & \color{blue}$\bm{0.78\pm0.10}$ & $\bm{0.77\pm0.12}$ & $0.40\pm0.07$ & $0.61\pm0.10$ & $0.34\pm0.07$ & $0.44\pm0.09$ & $0.46\pm0.09$ & $0.68\pm0.10$ & $0.45\pm0.09$ & $0.15\pm0.06$ & $0.32\pm0.07$ & $0.51\pm0.09$ \\ 
& Recall (PA) & $0.83\pm0.04$ & $0.73\pm0.03$ & $0.54\pm0.04$ & \color{blue}$\bm{0.96\pm0.02}$ & $0.75\pm0.05$ & $0.76\pm0.08$ & $0.81\pm0.07$ & $\bm{0.85\pm0.07}$ & $0.79\pm0.08$ & $0.80\pm0.08$ & $0.29\pm0.10$ & $0.78\pm0.08$ & $0.74\pm0.06$ \\ 
\midrule 
\multirow{4}{*}{Mean}& Accuracy & $\bm{0.88\pm0.01}$ & $0.86\pm0.01$ & $0.84\pm0.01$ & $0.86\pm0.01$ & \color{blue}$\bm{0.91\pm0.01}$ & $0.79\pm0.01$ & $0.84\pm0.01$ & $0.84\pm0.01$ & $\bm{0.88\pm0.01}$ & $0.85\pm0.01$ & $0.73\pm0.01$ & $0.81\pm0.01$ & -- \\ 
& F1 & \color{blue}$\bm{0.65\pm0.08}$ & $0.43\pm0.10$ & $0.34\pm0.09$ & $\bm{0.63\pm0.05}$ & \color{blue}$\bm{0.65\pm0.08}$ & $0.50\pm0.05$ & $0.58\pm0.06$ & $0.58\pm0.05$ & \color{blue}$\bm{0.65\pm0.08}$ & $0.55\pm0.06$ & $0.37\pm0.07$ & $0.52\pm0.09$ & -- \\ 
& Precision (UA) & $0.63\pm0.03$ & $0.63\pm0.10$ & $0.63\pm0.06$ & $0.56\pm0.02$ & \color{blue}$\bm{0.72\pm0.04}$ & $0.44\pm0.02$ & $0.51\pm0.02$ & $0.51\pm0.02$ & $\bm{0.68\pm0.03}$ & $0.51\pm0.02$ & $0.34\pm0.03$ & $0.47\pm0.03$ & -- \\ 
& Recall (PA) & $0.71\pm0.03$ & $0.38\pm0.03$ & $0.30\pm0.03$ & \color{blue}$\bm{0.81\pm0.02}$ & $0.65\pm0.02$ & $0.69\pm0.04$ & $\bm{0.74\pm0.04}$ & $\bm{0.74\pm0.04}$ & $0.68\pm0.04$ & $0.67\pm0.04$ & $0.55\pm0.04$ & $0.69\pm0.04$ & -- \\ 

\bottomrule
\end{tabular}}
\caption{Performance metrics and associated standard errors for each map and reference dataset compared in this study, including a majority vote ensemble of all 11 maps. The highest value in each row is in \textbf{\color{blue}bold blue} and the second-highest in \textbf{bold black}.}
\label{tab:metrics}
\end{table}

\subsection*{Map agreement}
We quantified the consensus between cropland classifications across all of the compared maps as well as the agreement between all pairs of maps. We characterized map agreement in three ways: 1) quantified as a percentage of total pixels with agreement across all maps, 2) illustrated visually as a map to show spatial patterns of agreement and disagreement, and 3) quantified as a percentage of total pixels that agree for pairwise maps as an agreement matrix.

\subsubsection*{Consensus across all maps}
Table \ref{tab:agreement} quantifies the agreement between crop and non-crop predictions made by all maps as a percentage of the total pixels in the 10 m/pixel resolution consensus map. The highest overall consensus, i.e., the highest percentage of all locations in which all maps predict the same class, was in Mali (69.9\%) and Kenya (60.6\%). The lowest overall consensus was in Rwanda (15.8\%) and Malawi (21.8\%). The \textit{crop consensus}, i.e., the percentage of locations in which all maps predict cropland, was very low---less than 0.5\% for all countries; the highest crop consensus was in Kenya (0.4\%) and Uganda (0.4\%). The \textit{non-crop consensus}, i.e., the percentage of locations in which all maps predict non-crop, was approximately equivalent to the overall consensus because non-crop predictions constituted the  majority of consensus pixels. This shows there is greater disagreement among maps in predicting where cropland \textit{is} compared to where cropland \textit{is not}. The ``split prediction'' column in Table \ref{tab:agreement} gives the percentage of pixel locations where the consensus between all maps is split; i.e., 5 or 6 maps predict crop while the remaining 6 or 5 maps predict non-crop. These are the regions of the lowest consensus across all maps. The split prediction percentage was highest in Rwanda (24.0\%) and lowest in Mali (2.9\%).

\begin{table}[ht]
\centering
\begin{tabular}{lcccc}
\toprule
Country & All predict same class & All predict crop & Split prediction & None predict crop (all non-crop)\\
\midrule
Kenya & 60.6 & 0.4 & 5.0 & 60.2 \\
Malawi & 21.8 & 0.0 & 17.5 & 21.8 \\
Mali & 69.9 & 0.0 & 2.9 & 69.9 \\
Rwanda & 15.8 & 0.1 & 24.0 & 15.8  \\
Tanzania & 44.4 & 0.0 & 9.0 & 44.4 \\
Togo & 23.8 & 0.2 & 17.0 & 23.6 \\
Uganda & 29.1 & 0.4 & 11.6 & 28.8 \\
Zambia & 49.5 & 0.0 & 4.4 & 49.5 \\
\hline
\end{tabular}
\caption{\label{tab:agreement}Agreement between 11 compared maps as percent of total pixels. ``Split prediction'' describes pixels where the prediction across all maps is approximately evenly split (5 predict crop but 6 predict non-crop, or 6 predict crop but 5 predict non-crop).} 
\end{table}

\subsubsection*{Spatial visualization of map agreement}
\begin{figure}[ht]
\centering
\includegraphics[width=\linewidth]{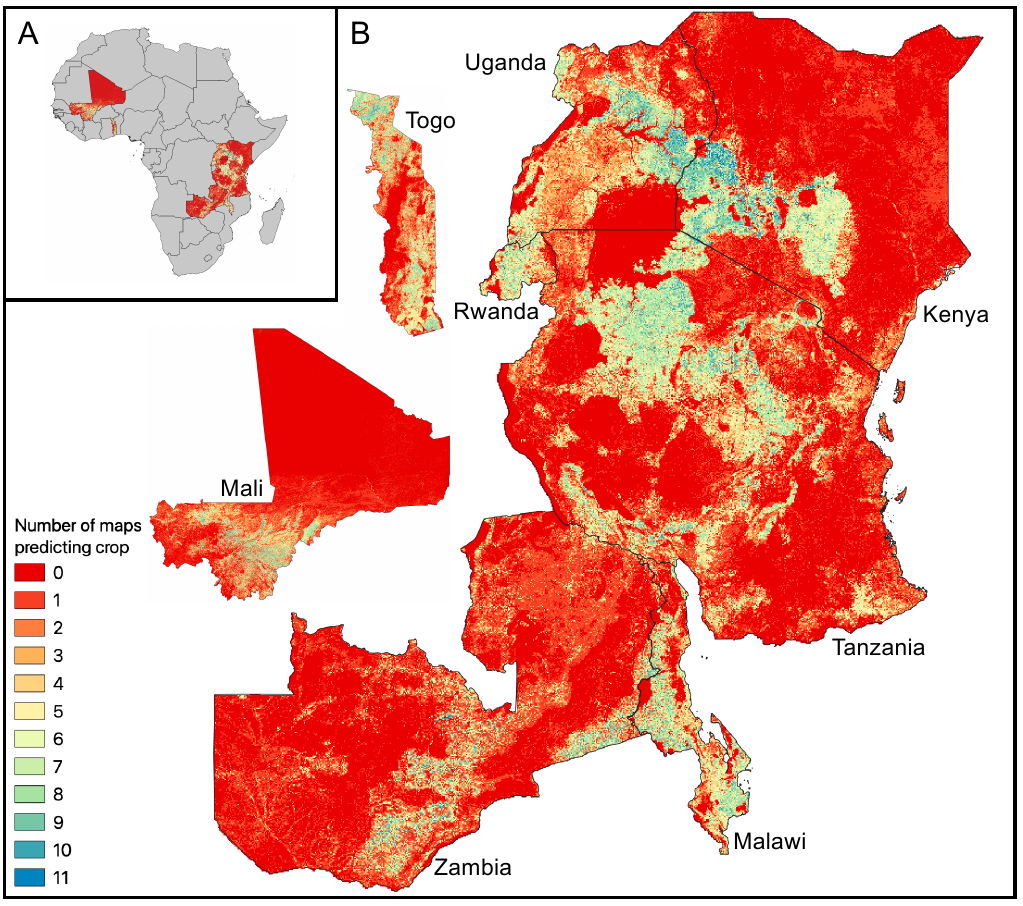}
\caption{Map showing the number of maps that predict cropland in each 10m/pixel location. Panel A shows the maps in the context of the African continent while Panel B gives larger views of the maps (countries not to scale). Blue and red indicate regions of high or unanimous agreement between maps for the crop and non-crop classes, respectively. Yellow indicates regions of high disagreement between maps. }
\label{fig:agreement}
\end{figure}

Figure \ref{fig:agreement} visualizes the spatial distribution of agreement between the 11 maps, where colors indicate the number of maps that predict the crop class in each 10m pixel location. Blue pixels (where the map value is 11) indicate locations where all maps unanimously or near-unanimously predict cropland (which constitutes less than 0.5\% of pixels in each country, as described in the previous section). Red pixels (where the map value is 0) indicate locations where all maps unanimously or near-unanimously predict the non-crop class. The fraction of unanimous non-crop pixels is especially high in Kenya, Zambia, Mali, and Tanzania, where the majority of the country's land area is not used for agriculture.\cite{faostat} Yellow pixels indicate locations of high disagreement between maps where approximately half of the maps predict crop and half predict non-crop. While some regions in each map appear primarily yellow, yellow-orange, or yellow-green (indicating homogeneous areas of disagreement), most of the disagreement regions are on the edges surrounding high agreement pixels of either class. This suggests that most zones of high disagreement lie on the boundaries between crop and non-crop regions, where transitions between land cover types might make discrete classification challenging. These zone are also likely locations of cropland expansion and are important to accurately identify for efficient resource allocation, crop rotation, pest management, environmental conservation, land use planning, and monitoring.

\subsubsection*{Pairwise map agreement}
\begin{figure}[ht]
\centering
\includegraphics[width=\linewidth]{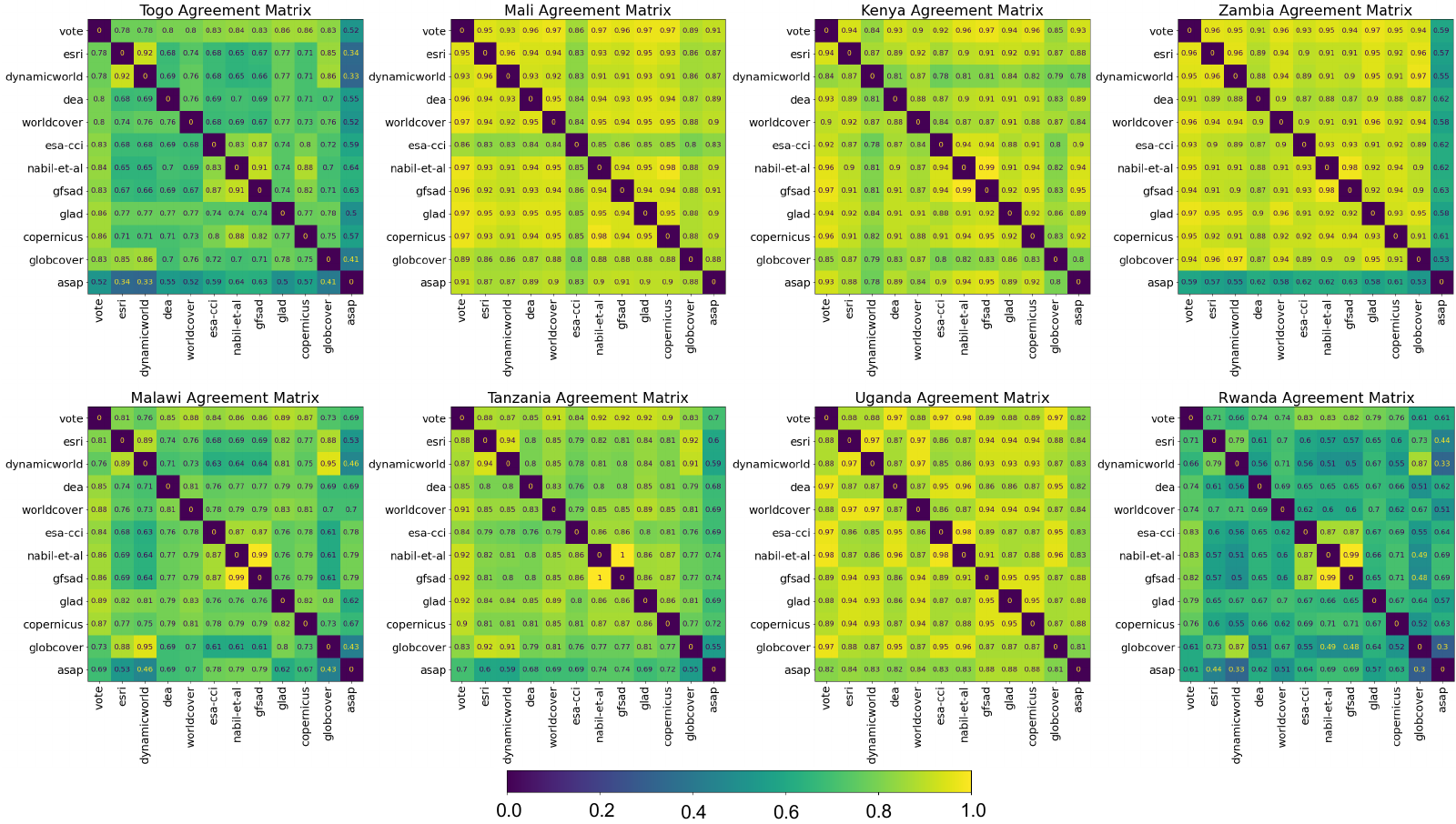}
\caption{Fraction of pixels that agree (predict same class) for each pair of compared maps, ordered by spatial resolution (high to low). Diagonal entries are set to 0. See online version for high resolution.}
\label{fig:agreement-matrix}
\end{figure}

\begin{figure}[ht]
\centering
\includegraphics[width=\linewidth]{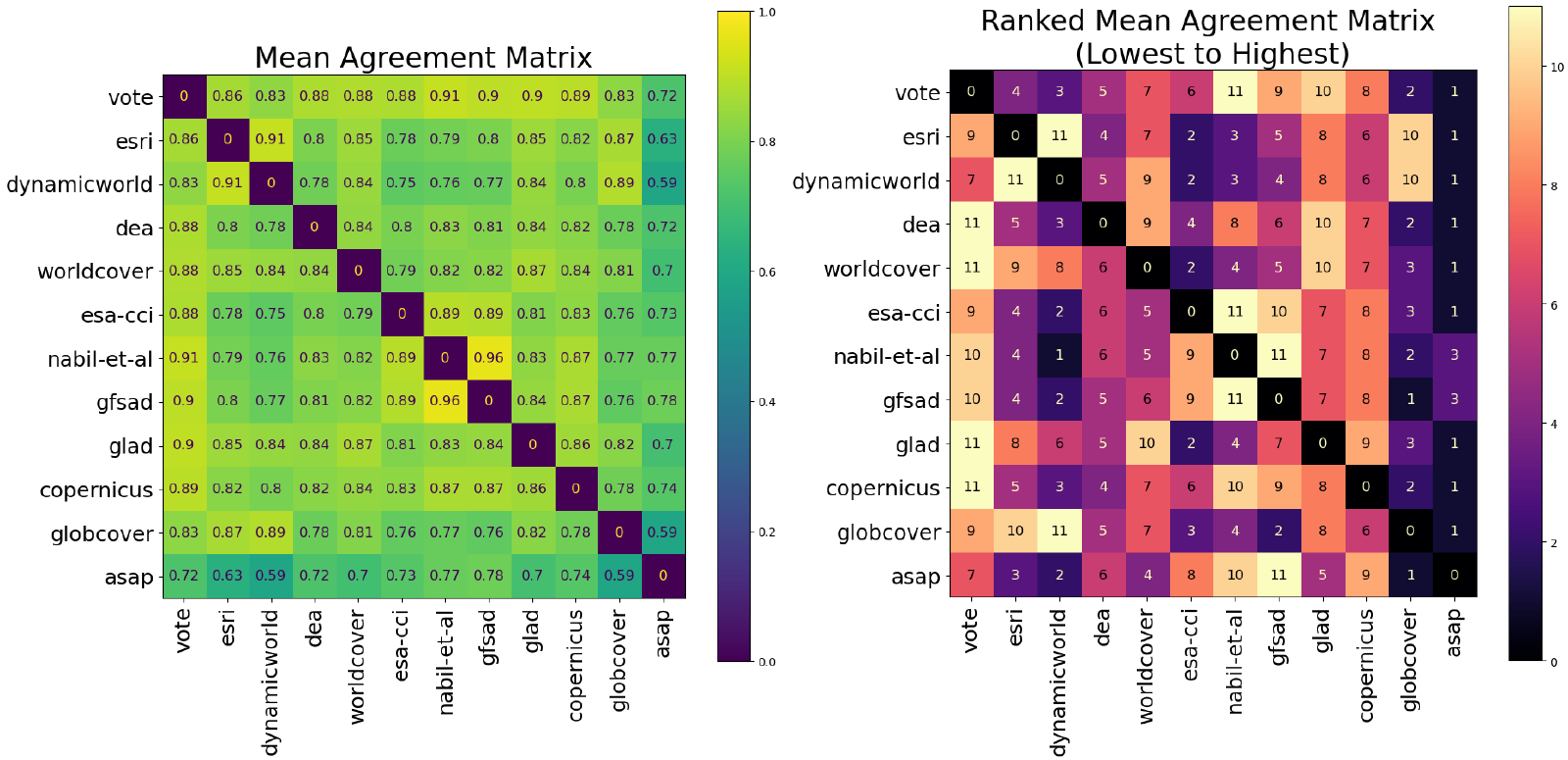}
\caption{Left matrix shows the average fraction of pixels that agree (predict same class) for each pair of compared maps, averaged over all countries (mean of matrices in Figure \ref{fig:agreement-matrix}). Right matrix shows the ranking of lowest (1) to highest (11) pairwise agreement between maps within each row. Diagonal entries are set to 0.}
\label{fig:agreement-matrix-mean}
\end{figure}

Figure \ref{fig:agreement-matrix} quantifies the pairwise agreement between each map compared in this study, and the majority vote ensemble map, using a symmetric agreement matrix for each country. Each cell reports the fraction of pixels that agree (predict the same class) for each pair of compared maps, ordered by spatial resolution (high to low). Since maps are ordered by spatial resolution (highest to coarsest, i.e., increasing ground sampling distance), blocky patterns about the diagonal indicate that maps with the same or similar resolution have the highest pairwise agreement. Diagonal cells are set to zero. Figure \ref{fig:agreement-matrix-mean} shows the mean agreement matrix (average agreement matrices for each country). All matrices use the same color bar ranging from 0 to 1, so relative agreement fractions can be compared across all countries. The pairwise agreement between maps is low for Rwanda, Togo, and Malawi and relatively high for Zambia and Mali. Overall, the maps that share the highest agreement are Nabil et al. and GFSAD, Nabil et al. and majority vote, and Dynamic World and Esri. The Nabil et al. map has high agreement with the majority vote map with GFSAD because it combines the predictions of GFSAD and two other maps used in the study.

Figure \ref{fig:agreement-matrix-mean} shows the order of pairwise maps ranked from lowest to the highest mean pairwise agreement within each row. There is high agreement between Nabil et al.\cite{nabil2022constructing} and the ESA-CCI, Copernicus, and GFSAD crop masks, which is expected since the Nabil et al. mask is a combination of these maps. ASAP has the overall lowest pairwise agreement with other maps, and ESA-CCI has particularly low agreement with all other maps in Mali. Surprisingly, GlobCover agrees most with Dynamic World and Esri. Digital Earth Africa, WorldCover, GLAD, and Copernicus all agree most with the majority vote map, meaning they are most consistent with the majority prediction across all the maps. Excluding the majority vote, GLAD agrees most with WorldCover, and vice versa, and Digital Earth Africa agrees most with GLAD. 

\subsection*{Accuracy vs. spatial resolution}
\begin{figure}[ht]
\centering
\includegraphics[width=\linewidth]{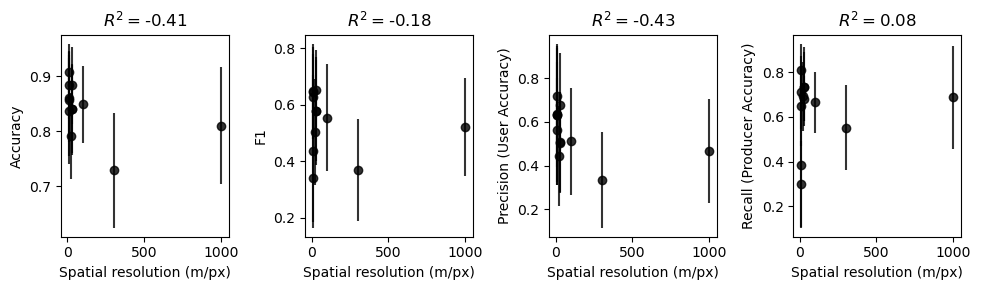}
\caption{Scatter plots showing mean performance metrics (averaged across all countries) vs. spatial resolution (in m/pixel). }
\label{fig:perf-vs-res}
\end{figure}

Previous work has hypothesized that land cover products based on higher resolution satellite observations (e.g., 10 m/pixel Sentinel-2) should have improved cropland mapping performance in Sub-Saharan Africa as a result of small and fragmented field sizes being more clear in the higher-resolution images,\cite{perez2017comparison} though other work found that small field sizes were not the primary influencing factor on the disagreement between maps.\cite{nabil2020assessing} Figure \ref{fig:perf-vs-res} shows a scatter plot of the performance metric for each map, averaged across all countries, vs. spatial resolution (in m/px). The title of each plot reports the Pearson product-moment correlation coefficient. 

Accuracy and precision (user's accuracy) show a moderate negative correlation with spatial resolution (R$^2$ of -0.41 and -0.43, respectively); i.e., accuracy and precision scores tend to be higher for higher resolution maps. However, we found no correlation between recall (producer's accuracy) and spatial resolution (R$^2$=0.08). We found a weak negative correlation between F1 score and spatial resolution (R$^2$=-0.18). This suggests that finer spatial resolution satellite data may help reduce false positives (other land cover types incorrectly classified as cropland) but not false negatives (cropland incorrectly classified as other land cover).

When interpreting the correlation between spatial resolution and performance metrics, it is important to consider the impact of the ASAP and GlobCover maps, which have substantially coarser spatial resolution than the other maps. ASAP has a resolution of 1000 m/px and GlobCover has a resolution of 300 m/px, while the remaining maps range from 10-100 m/px. After excluding ASAP and GlobCover, the correlation between spatial resolution and each metric was R$^2$ of -0.26 for accuracy, 0.26 for F1, -0.43 for precision, and 0.42 for recall. Thus, for maps with spatial resolution of 100 m/px and below, we found a weak to moderate \textit{positive} correlation between F1 and recall and spatial resolution, meaning coarser spatial resolution products tend to have higher F1 and recall scores. However, we found a weak to moderate \textit{negative} correlation between accuracy and precision, meaning finer spatial resolution products tend to have higher accuracy and precision. The intended use case of the map should influence which metrics should be weighted most highly in choosing the most suitable map.
 
\subsection*{Accuracy vs. temporal mismatch}
\begin{figure}[ht]
\centering
\includegraphics[width=\linewidth]{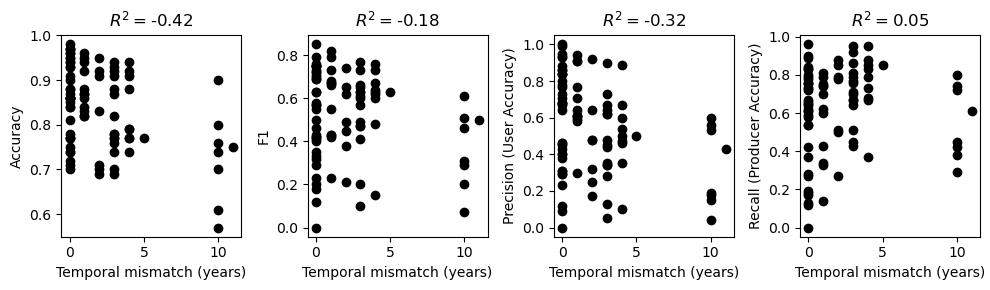}
\caption{Scatter plots showing performance metrics vs. temporal mismatch between the map year and the reference data year (unit is the number of years).}
\label{fig:perf-vs-temp-mismatch}
\end{figure}

Previous work has stressed the importance of using a crop mask with the same (or as close as possible to the same) year as the downstream analysis, e.g., when assessing in-season crop conditions.\cite{whitcraft2019no} Figure \ref{fig:perf-vs-temp-mismatch} plots each metric against the temporal mismatch (number of years difference) between the reference data year and the map year (calculated as the absolute difference between the map year and the reference data year). We found a weak to moderate negative correlation between accuracy, F1, precision, and temporal mismatch (R$^2$ of -0.42, -0.18, and -0.32, respectively), indicating that higher performance somewhat correlates with closer temporal matching. We found no correlation (R$^2$ of 0.05) between recall (producer's accuracy) and temporal mismatch. 

As in the case of spatial resolution, it is important to consider the effect of outliers in this assessment. The nominal year for the GlobCover map is 2009 and is an outlier among the map years, which otherwise range from 2017 to 2020. When GlobCover is excluded, the R$^2$ between temporal mismatch and each metric is -0.22 for accuracy, 0.06 for F1, -0.22 for precision, and 0.30 for recall. In other words, for maps within 5 years of the reference data year, there is a weaker negative correlation between temporal mismatch and accuracy, F1, and precision. However, there is also a weak positive correlation between temporal mismatch and recall. These results suggest that using maps within a few years of the reference data or analysis year may be sufficient for most studies, and an exact temporal match may not be needed for all analyses. The importance of temporal matching between map year and reference/analysis year should depend on the intended use of the map. 

\subsection*{Impact of map differences in downstream use}
Cropland maps are commonly used as masks to select the cropland pixels from satellite datasets for crop conditions assessments, yield estimation, and other downstream analyses.\cite{whitcraft2019no} For example, in the Global Agricultural Monitoring (GLAM) System, a user can assess current season crop conditions by comparing the normalized difference vegetation index (NDVI) time series of the current year (a measure of vegetation condition throughout the year) to that of a typical year or range of years (representing average vegetation conditions).\cite{becker2010monitoring} 

\begin{figure}[ht]
\centering
\includegraphics[width=0.8\linewidth]{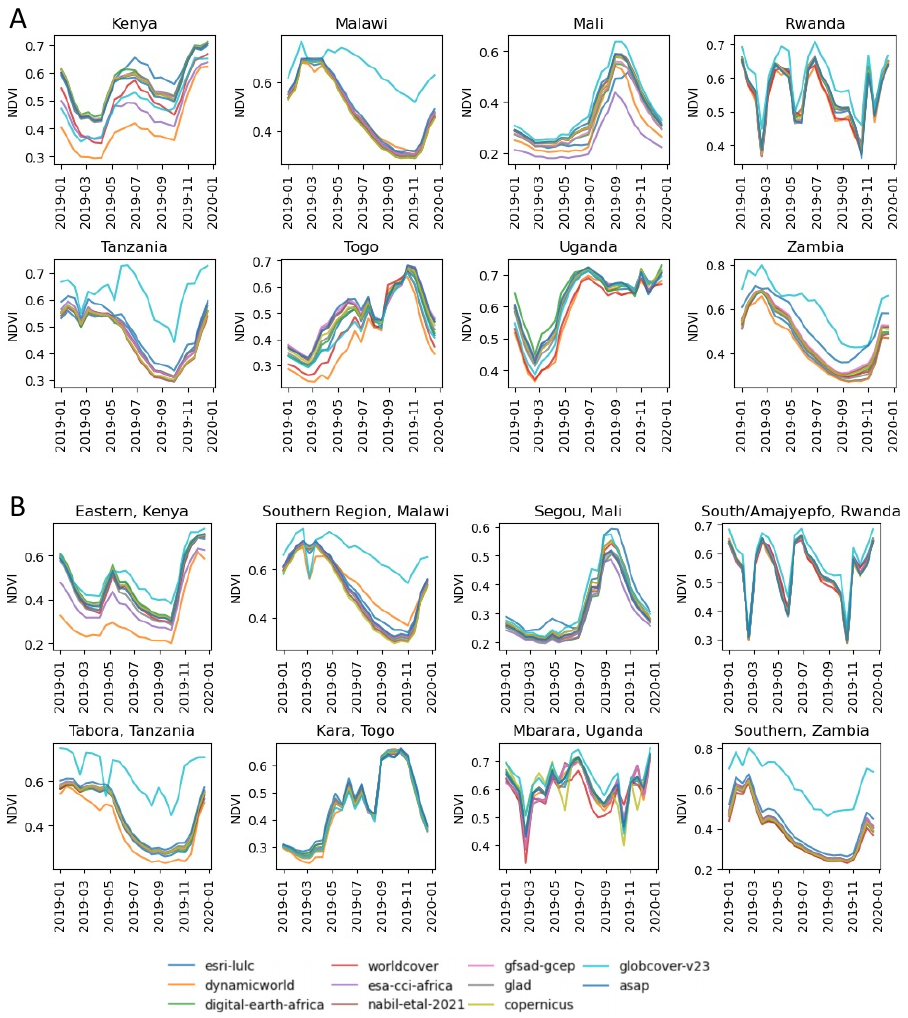}
\caption{Time series of 16-day MODIS MOD13A1.061 Terra Vegetation Indices averaged within (A) each country and (B) an administrative level 1 region, masked using each crop mask assessed in this study.}
\label{fig:ndvi}
\end{figure}

In Figure \ref{fig:ndvi}, we illustrate how the choice of crop mask can affect the NDVI time series and, thus, the end-user's interpretation of vegetation conditions. Each plot shows a time series of the mean NDVI from the MODIS MOD13A1.061 Terra Vegetation Indices 16-Day Global 500m data product, created by masking the image collection using each map and computing the mean of all pixels within the country boundary (panel A) and an administrative level 1 (admin1) region (panel B). Admin1 refers to the largest subnational administrative unit within a country, e.g., states in the United States.\cite{faogaul} We chose specific admin1 regions based on areas of high disagreement between maps to highlight discrepancies that may arise due to map differences. These plots reveal that GlobCover substantially overestimates crop conditions as measured by NDVI compared to other maps, particularly in Malawi, Tanzania, Rwanda, and Zambia. In some countries such as Malawi, Tanzania, and Zambia, most masks result in similar time series, but there is a large spread in others such as Kenya, Mali, and Togo. Notably, Kenya had the highest percentage of pixels in agreement across all maps of the 8 countries (Table \ref{tab:agreement}), but the greatest differences between the maps when used as crop masks for vegetation indices. 

These results highlight the importance of considering the locations and other characteristics of discrepancies between maps and not just the magnitude of those discrepancies in future work. We used the MODIS data product since this is commonly used by current agriculture monitoring systems such as GLAM.\cite{becker2010monitoring} Since the MODIS NDVI data product used has 500 m/pixel resolution, the time series are aggregated using each crop mask resampled to 500 m/pixel resolution to match the NDVI data, thus most of the masks are downscaled to a much coarser resolution than their true resolution. If a higher-resolution NDVI data product (such as Landsat or Sentinel-2) were used, we expect the time series resulting from each mask would diverge even more substantially than in the downscaled MODIS time series. 

\section*{Discussion}

\subsection*{Which map is best?}
No single map can be considered optimal across all evaluations, metrics, and countries, thereby leaving users of these maps uncertain about the most suitable choice for a given use case or application. The WorldCover and GLAD maps had good performance for most countries in the reference dataset evaluation, high agreement with the majority vote, and similar NDVI time series to the majority of maps. GlobCover had low performance for most countries in the reference dataset evaluation, moderate agreement with the majority vote, and NDVI time series that differed substantially from the majority of maps. Thus WorldCover or GLAD are likely to be reasonable choices for many use cases, while GlobCover should be used cautiously. 

The best map for a given use case should be chosen based on the performance and relevance of that map for the specific use case. Users should identify the right map based on the region-specific evaluation data or their intended use case. This means that users should develop use case-specific evaluation datasets and experiments for choosing the best map for their intended use case if relevant datasets do not exist already.  Users should also visually inspect candidate maps in their intended study area to qualitatively assess map suitability, as quantitative metrics may not fully capture map performance for a given use case.\cite{rolf2023evaluation}

\subsection*{Regional vs. global models}
Researchers have been shifting toward training single, global models that are optimized to predict land cover anywhere in the world. Global models compared in this study are Esri, Dynamic World, and WorldCover. Global models are appealing because they simplify computation and enable models to learn from more diverse data, but these benefits may come at the expense of high intra-class variance that makes learning to predict classes accurately across a wide range of agro-ecological conditions challenging. In contrast, regional models are trained to optimize classification performance for a regional sub-group, such as an agro-ecological zone, a global grid tile, or a country (see Table \ref{tab:products-evaluated}). Table \ref{tab:metrics} shows that models trained to optimize regional performance tend to outperform global models. In addition to training a global model, the Esri and Dynamic World maps are the only models that do not include temporal features in the input. Classification is based on segmentation of a single image (or composite image). Thus, spatial and spectral information is available to the model but not temporal. Temporal information is important for identifying cropland and could be another factor contributing to better performance of some maps over others.

\subsection*{Ensemble models}
Since ensembles often perform better than individual base classifiers, it may be surprising that the Majority Vote ensemble does not give the best overall performance. However, necessary conditions for an ensemble of base classifiers to perform better than individual base classifiers are that the base classifiers are independent of each other (make independent errors) and perform better than random guessing,\cite{tan2007introduction} which is not necessarily true for all of the maps in this study. The Nabil et al.\cite{nabil2022constructing} crop mask is also an ensemble; it combines four data products, three of which are also compared in this study (GFSAD, ESA-CCI, and Copernicus). These maps are highly correlated (see Figure \ref{fig:agreement-matrix}). Nabil et al. proposed the ensemble crop mask to provide a combined product better than the individual maps. The Nabil et al.\cite{nabil2022constructing} ensemble mask performs better overall than GFSAD, ESA-CCI, and Copernicus individually but is not among the highest-performing maps of all those compared in this study.

\subsection*{Future efforts}
This study's results show low consensus between the 11 compared maps in each of the studied countries, particularly for the cropland class. Unanimous agreement between maps on cropland locations is rare. In addition, the average performance across all of the maps in each country is quite low. Most previous efforts to create more accurate global land cover maps focus on creating new methods and models that optimize performance over samples drawn globally. We encourage future work focusing on targeted improvements that boost performance for the lowest-performing sub-groups (e.g., countries or biomes) among existing methods. This would help to resolve inconsistencies between existing maps and reduce the disparity in performance across countries globally, resulting in more geographically fair land cover maps and associated machine learning models. The results from this study can be used to inform such future work. For example, future efforts to create labeled datasets could focus on low-performing countries (e.g., Mali) or collect samples from regions of high disagreement between maps using random uniform sampling or stratified sampling in which consensus levels define strata. Future work could extend this analysis for more countries and maps using our publicly provided code. Future work could also help direct efforts to improve map accuracy and provide useful information to map users by identifying the types of land cover that are commonly confused with cropland in these maps. 

We hope that this study's reference datasets and results can provide a useful benchmark for evaluating future map products against a common baseline. We provide our reference datasets to enable this benchmarking and for transparency but emphasize that future inter-comparison with these results requires researchers to use the reference datasets only for final, independent accuracy assessment and not use them during the model training or map creation process.  

\section*{Methods}
We designed our accuracy assessment to meet the good practice criteria described in Stehman \& Foody (2019):\cite{stehman2019key} 
\begin{enumerate}
    \item Map relevant: the accuracy estimates and error matrices reflect the proportional area representation of the study region via uniform random sampling and we report unnormalized error matrices in terms of sample counts and population error matrices in terms of map area proportion in Supplementary File 1.
    \item Statistically rigorous: we implemented a probability sampling design of simple uniform or stratified random sampling for each country. We quantified the variability of the accuracy by reporting standard errors. 
    \item Quality assured: we established protocols to monitor and evaluate the quality of reference data of results including assigning two or more interpreters to label each point while blind to the map category; only points with unanimous interpreter consensus were used in the analysis.
    \item Reliable: The variability among interpretations of reference sample labels is very low since we only used points that had unanimous labeler agreement in analysis. It is likely that different samples and different interpreters (with the same training) would lead to similar results. 
    \item Transparent: We have provided all relevant details to inform readers about the quality of the results.
    \item Reproducible: We have provided the reference sample datasets used in the analysis in a public Google Earth Engine asset and Zenodo repository. We provided the code used to preprocess and evaluate all maps in a GitHub repository. The code for generating the samples and instructions provided to interpreters is available upon request.
\end{enumerate}

\subsection*{Cropland and Land Cover Maps}
We analyzed 11 global and continent-scale land cover and land use maps made publicly available in recent years. Several of these maps have not been included in previous studies comparing publicly available land cover maps. These maps are summarized in Table \ref{tab:products-evaluated}, which span a range of temporal availability, spatial resolutions, and geographic coverage. Table \ref{tab:products-evaluated} provides the cropland definition specified by each data product. Each map was created by making dense, contiguous predictions over a geographic area using a trained machine learning classifier (though the classifier used for each differs). 

The following subsection headings categorize these map products as being produced using \textit{regional} or \textit{global} models. In the regional model category, separate machine learning models are optimized for a specific region (e.g., a country, continent, or agroecological zone) that generates a predicted map for that region. These regional maps may then be combined to cover a larger geographic area, such as a global map product. In the global model category, a single machine learning model is optimized for making predictions in all regions of the world, where data used for training may be aggregated from many different regions. A single global model is then used to make predictions everywhere to form one predicted global map. 

The values of each map at the reference sample locations were obtained using the \texttt{reduceRegions()} function in Google Earth Engine sampled at the native resolution of each map. This value extraction is fully reproducible using the code provided in the Code Availability section. All maps were clipped to the border of each country in the study and converted to a binary crop/non-crop map. Consensus and agreement analyses were performed after resampling all maps to a common spatial resolution of 10 m/px. In the subsequent sections, we summarize how each map was created and used in this study. 

\begin{table}[ht]
\resizebox{\columnwidth}{!}{%
\renewcommand{\arraystretch}{0.3}
\begin{tabular}{p{0.2\linewidth}p{0.1\linewidth}p{0.05\linewidth}p{0.06\linewidth}p{0.1\linewidth}p{0.5\linewidth}}
\toprule \\
Dataset & Year(s) & Res. (m/px) & Model Scale & Coverage & Definition of Cropland\\
\midrule
DEA Cropland Extent & 2019 & 10 & AEZ & Continent & ``A piece of land of minimum 0.01 ha that is sowed/planted and harvestable at least once within the 12 months after the sowing/planting date.''\cite{burton2022co,dea-data} \\ \\

Dynamic World & 2015-2023 & 10 & Global & Global & ``Human planted/plotted cereals, grasses, and crops''\cite{brown2022dynamic}\\ \\

Esri LULC & 2017-2022 & 10 & Global & Global & ``Human planted/plotted cereals, grasses, and crops not at tree height; examples: corn, wheat, soy, fallow plots of structured land.'' \cite{karra2021global,esri-data}
\\ \\

ESA WorldCover & 2020-2021 & 10 & Global & Global & ``Land covered with annual cropland that is sowed/planted and harvestable at least once within the 12 months after the sowing/planting date. The annual cropland produces an herbaceous cover and is sometimes combined with some tree or woody vegetation. Note that perennial woody crops will be classified as the appropriate tree cover or shrub land cover type. Greenhouses are considered as built-up.''\cite{zanaga2022esa,worldcover-data, worldcover-manual}  \\\\ %

ESA CCI Africa & 2016 & 20 & Continent & Continent & No explicit definition provided.\cite{cci_africa_feedback,cci-data} \\\\ %
GFSAD & 2015 & 30 & AEZ & Global & ``Cropland that is cultivated and harvested for food, feed, and (or) fiber, one or more times during a 12-month period; Cropland that is left fallow, even when equipped for agriculture; and 
 Cropland that is permanently cropped with plantations (for example, orchards, vineyards, coffee, tea, and rubber).''\cite{thenkabail2021global,gfsad-data} \\ \\ %
Nabil et al. & 2016 & 30 & Mixed & Continent & ``all agricultural annual standing croplands, cropland fallows, and permanent plantation crops''\cite{nabil2022constructing,nabil-data} \\ \\
GLAD & 2003, 2007,2011, 2015, 2019   & 30 & $1^{\circ}\times1^{\circ}$ & Global & ``[...] land used for annual and perennial herbaceous crops for human consumption, forage (including hay) and biofuel. Perennial woody crops, permanent pastures and shifting cultivation are excluded from the definition.'' \cite{potapov2022global,glad-data}\\\\ %
Copernicus Land Cover & 2015-2019 & 100 & Biome & Global & ``Cultivated and managed vegetation / agriculture. Lands covered with temporary crops followed by harvest and a bare soil period (e.g., single and multiple cropping systems). Note that perennial woody crops will be classified as the appropriate forest or shrub land cover type.''\cite{buchhorn2020copernicus,cgls-data}\\\\
ESA GlobCover & 2005, 2009 & 300 & Strata & Global & ``Post-flooding or irrigated croplands,'' ``rainfed croplands,'' ``Mosaic Cropland (50-70\%) / Vegetation (grassland, shrubland, forest) (20-50\%),'' and ``Mosaic Vegetation (grassland, shrubland, forest) (50-70\%) / Cropland (20-50\%)''\cite{teamglobcover,globcover-data} \\ \\
ASAP Crop Mask & 2017 & 1000 & Mixed & Global & ``arable land and permanent crops...independently of their life forms (e.g., tree forms), production systems (i.e., both rainfed and irrigated), and density of cover''\cite{perez2017comparison,rembold2019asap,asap-data}\\\\ %
\bottomrule
\end{tabular}%
}
\caption{\label{tab:products-evaluated}Description of map data products evaluated in this study.}
\end{table}

\subsubsection*{Regional models}

\paragraph{Digital Earth Africa.} The Digital Earth Africa (DEA) Cropland Extent map\cite{burton2022co} estimates crop extent at 10 m/pixel resolution for the year 2019 in Africa. The continent-scale map was created by combining maps predicted for eight different agro-ecological zones (AEZ) within the African continent covering Eastern, Western, Northern, Sahel, Southern, Southeast, Central, and Indian Ocean regions of Africa. A separate random forest model was trained using data sampled from each AEZ and used to predict a complete map for that AEZ. The input feature vector contains hand-crafted features (temporal statistics and geomedian composites) from a 12-month time series of Sentinel-2 multispectral observations and ancillary datasets (topography and climatology). All maps and associated training and validation datasets are for 2019 with 10 m/pixel spatial resolution. Digital Earth Africa reported an overall accuracy of 90.3\% in Eastern Africa (includes Kenya), 83.6\% in Western Africa (includes Togo), and 87.3\% in Southeast Africa (includes Malawi). We accessed the DEA Cropland Extent map using the following GEE asset: \texttt{ee.ImageCollection("projects/sat-io/open-datasets/DEAF/CROPLAND-EXTENT/mask")} .

\paragraph{ESA CCI Land Cover Africa.} This 20 m/pixel land cover map of Africa for the year 2016 was produced by UC Louvain as part of the European Space Agency (ESA) Climate Change Initiative (CCI). The classification into 10 land cover classes (including ``cropland'') is based on input features extracted from one year (December 2015-December 2016) of Sentinel-2A multispectral observations. The product documentation states that classification was performed using ``two classification algorithms, the Random Forest (RF) and Machine Learning (ML)'' but does not specify which machine learning technique is used in the second algorithm.\cite{cci_africa_feedback} Independent evaluations of the map accuracy reported overall accuracy of 56\% in Kenya\cite{lacowiki2020} and around 65\% in over the continent, noting that the map should be improved before being used for research or practical purposes.\cite{lesiv2017evaluation} We binarized the dataset by converting pixels labeled with the cropland class to labels of 1 and 0 otherwise.
We accessed the CCI Land Cover Africa map using the following GEE asset: \texttt{ee.Image("projects/sat-io/open-datasets/ESA/ESACCI-LC-L4-LC10-Map-20m-P1Y-2016-v10")}.

\paragraph{GFSAD Global Cropland Maps.} The goal of the Global Food Security Support Analysis Data (GFSAD) project was to create a global map of cropland extent at 30 m/pixel resolution based on Landsat satellite Earth observation data.\cite{thenkabail2021global} The GFSAD Global Cropland Extent Product (GCEP) was created by combining maps predicted for 74 AEZs defined globally. A predicted map was created for each AEZ by combining the outputs of four machine learning models---random forest, support vector machine (SVM), an object-based classifier, and recursive hierarchical segmentation---trained for binary cropland classification. The input features were derived from Landsat-8 multispectral time series observations in addition to elevation and slope attributes derived from the SRTM DEM for years 2013-2016, with outputs intended to represent the nominal year 2015. The study reported overall accuracy of 93.7\% in Africa and, more specifically, 91.3\% in AEZ 36 (includes Kenya and Malawi) and 90.8\% in AEZ 34 (includes Togo).
We accessed the GFSAD map using the following GEE asset: \texttt{ee.ImageCollection("projects/sat-io/open-datasets/GFSAD/GCEP30")}.

\paragraph{GLAD.} The GLAD Global Cropland Maps provide binary cropland classifications at 30 m/pixel for 2003, 2007, 2011, 2015, and 2019. Classification is performed using bagged decision trees with features extracted from time series of Landsat Analysis Ready Data (ARD).\cite{potapov2022global} A separate model is trained for each $1^\circ\times1^\circ$ ARD tile, and the predictions from each tile are merged to form a global map. The regional accuracy reported for Africa for 2016-2019 was $96.5 \pm 0.8$, and the global map was noted to underestimate the cropland area in Africa due to the spatial resolution limitations. We binarized the 2019 dataset by converting pixels with values $>0.5$ to a 1 and otherwise 0 label.
We accessed the GLAD map using the following GEE asset: \texttt{ee.ImageCollection("users/potapovpeter/Global\_cropland\_2019")}.

\paragraph{ESA GlobCover.} The ESA GlobCover project aimed to provide global land cover maps based on observations from the 300 m/pixel MERIS satellite sensor. The project produced global land cover maps for 2009 (v2.3, based on observations from January-December 2009) and 2005 (v2.2, based on observations from December 2004-June 2006); we used v2.3 in this study. A separate classifier was trained for each of the 22 global strata designed to reduce land surface reflectance variability in the data processed by each regional classifier. Classification into 22 land cover classes is achieved through a four-stage process of (1a) supervised classification of under-represented classes, (1b) unsupervised clustering of pixels not classified in step 1a, (2) temporal characterization of spectral clusters from step 1b, (3) aggregation of spectral clusters into fewer spectro-temporal clusters based on similar temporal patterns, and (4) rule-based classification of spectro-temporal clusters.\cite{teamglobcover} Defourny et al. \cite{defourny2009} reported an overall accuracy of 73\% but did not report Africa-specific metrics.\cite{defourny2009} We binarized the dataset by converting pixels labeled with the ``post-flooding or irrigated croplands'', ``rainfed croplands'', and ``mosaic cropland (50-70\%) / vegetation (grassland, shrubland, forest) (20-50\%)`` classes as 1, and all other classes as 0.
We accessed the ESA GlobCover map using the following GEE asset: \texttt{ee.Image("ESA/GLOBCOVER\_L4\_200901\_200912\_V2\_3")}.

\paragraph{Copernicus Land Cover.} The Copernicus Global Land Service Land Cover (CGLS-LC) map provides global land cover maps annually from 2015-2019 based on observations from the 100 m/pixel PROBA-V satellite sensor.\cite{buchhorn2020copernicus} The land cover classification includes 23 classes based on the United Nations Food and Agriculture Organization (UN-FAO) Land Cover Classification System and surface area statistics for 10 land cover types. More than 100 input features, including vegetation indices and time series statistics, were extracted from the PROBA-V multispectral satellite observations for 141,000 training sample locations. These samples were used to train a random forest classifier separately for each biome, where biomes were clusters of pixels determined from multiple global ecological datasets. Tsendbazar et al. (2020) reported a precision score (user's accuracy) of $62.4\pm3.7\%$ and a recall score (producer's accuracy) of $57.3\pm3.6\%$ for the cropland class in Africa and $80.1\pm 2.0\%$ overall accuracy in Africa.\cite{tsendbazar2020copernicus} We used the discrete classification in the 2019 map and created a binary cropland map by converting pixels labeled with the ``Cultivated and managed vegetation/agriculture'' class to 1 and all other classes to 0.
We accessed the Copernicus Land Cover map using the following GEE asset: \texttt{ee.ImageCollection("COPERNICUS/Landcover/100m/Proba-V-C3/Global")}.

\subsubsection*{Global models}

\paragraph{Esri Land Use/Land Cover.} The Esri Land Use and Land Cover (LULC) map, created by Esri, Impact Observatory, and Microsoft, provides a global classification of 9 land cover and land use classes (including crops) at 10 m/pixel resolution for the years 2017-2021.\cite{karra2021global} Classification was performed using a U-Net deep convolutional neural network model for semantic segmentation. The U-Net was trained using an extremely large dataset of over 5 billion human-labeled pixels paired with Sentinel-2 multispectral image composites. To create a LULC map for a given year, the least cloudy scenes over the year are selected and predictions are made for every Sentinel-2 tile, then combined by taking a class-weighted mode of all predictions. Karra et al. (2021) reported an overall accuracy of 85\% but did not provide more granular metrics for countries in Africa. We binarized the dataset by converting pixels labeled with the crop class to a label of 1, and otherwise 0.
We accessed the Esri LULC map using the following GEE asset: \texttt{ee.ImageCollection("projects/sat-io/open-datasets/landcover/ESRI\_Global-LULC\_10m\_TS")}.

\paragraph{ESA WorldCover.} The ESA WorldCover global land cover product was designed to build on the lessons learned from the ESA GlobCover and ESA CCI Land Cover products described in the previous section. WorldCover provides a global land cover classification map at 10 m/pixel resolution for 11 classes (including cropland).\cite{worldcover-manual} A large number of input features were extracted from Sentinel-2 multispectral, Sentinel-1 synthetic aperture radar, and several other Earth observations datasets (similar to the features extracted for the Copernicus Land Cover map described previously), in addition to localizing features such as the latitude/longitude position. These features were then used to train a CatBoost decision tree classifier, the outputs from which are post-processed using expert rules designed to reduce classification errors. The product validation report reported a precision score (user's accuracy) of $71.4\pm0.7\%$ and recall score (producer's accuracy) of $50.8\pm0.7\%$ for the cropland class in Africa and an overall accuracy of $73.6\pm0.2\%$ in Africa.\cite{worldcover-validation} WorldCover provides a v100 map for the year 2020 and a v200 map for the year 2021; we used v100 for year 2020, which most closely matches our reference data. 
We accessed the ESA WorldCover map using the following GEE asset: \texttt{ee.ImageCollection("ESA/WorldCover/v100")}.

\paragraph{Dynamic World.} Unlike the previous cropland and land cover maps that represent the land cover status of map locations based on satellite observations collected over a year or multiple years, Dynamic World represents the land cover status at a particular \textit{date}. Dynamic World provides a new global land cover classification at 10 m/pixel for every new cloud-free Sentinel-2 observation, which is nominally acquired every 5 days everywhere on Earth.\cite{brown2022dynamic} Dynamic World uses a fully-convolutional neural network (FCNN) to segment (i.e., classify all pixels in an input image) the land cover classes in a given Sentinel-2 multispectral image, trained on globally-distributed samples. Dynamic World provides land cover class probabilities and class labels (the class with the highest probability in each pixel) for nine land cover classes, including cropland. While Dynamic World is meant to capture near real-time land cover, it can represent a longer time period by computing a mode composite over that time period (e.g., one year as in the other maps in this study). We used a mode composite of Dynamic World land cover classes corresponding to the reference data year of each test region (e.g., 01/01/2019-12/31/2020 for Kenya) and then binarized the map so that the crop class had a label of 1 and all other classes had a label of 0 (representing non-crop classes). Brown et al. (2022) reported an accuracy of 88.9\% for the crop class, noting that Dynamic World tended to identify crops more poorly than other classes.\cite{brown2022dynamic} 
We accessed the Dynamic World map using the following GEE asset: \texttt{ee.ImageCollection("GOOGLE/DYNAMICWORLD/V1")}.

\subsubsection*{Ensemble maps}

\paragraph{ASAP Crop Mask.} The Anomaly hot Spots of Agricultural Production (ASAP) crop mask was created to provide a mask that could be applied to various Earth observation datasets to provide timely information about potential crop production anomalies using metrics computed from earth observations in crop areas.\cite{rembold2019asap} The ASAP global crop mask combines land cover and land use maps from multiple sources. For Africa, ASAP used the 250 m/pixel land cover map from Vancutsem et al. (2013), which combined ten different regional and global land cover maps based on expert judgment to create a ``best available'' crop mask for Africa. In certain countries for which a national crop mask was available (Afghanistan, Argentina, Australia, Europe, Mexico, and USA), that map was used in ASAP. In all other countries, the GlobCover\cite{teamglobcover} classification was used. All datasets were resampled to 1 km/pixel resolution with pixel values interpreted as the crop area fraction in each pixel (ranging from 0 to 100\%). We re-classified pixels with crop area fraction ranging from 5-95\% to 1, and all other pixels to 0, following the procedure described in user documentation (\url{https://glam1.gsfc.nasa.gov/api/doc/cropmask/v1/EC-JRC-ASAP-LC_v02_crops}).
We accessed the ASAP Crop Mask using the following GEE asset: \texttt{ee.Image("projects/sat-io/open-datasets/landcover/AF\_Cropland\_mask\_30m\_2016\_v3")}

\paragraph{Nabil et al. (2022).}\cite{nabil2022constructing} Similar to the ASAP crop mask, the combined crop mask published by Nabil et al. (2022) combines the best of four compared land cover maps within each of 41 AEZs to create a more accurate cropland map for the African continent than each of the individual maps.\cite{nabil2022constructing} The authors compared the accuracy in each AEZ of four maps---ESA-S2-LC20,\cite{cci_africa_feedback} GFSAD Global Cropland Extent,\cite{thenkabail2021global} CGLS-LC100-2016,\cite{buchhorn2020copernicus} and FROM-GLC30-2017\cite{feng2018multiple} (the first three of which we included in this study)---using a combination of labeled validation datasets. For each AEZ, the compared map that had the highest accuracy was used in the final map. The majority of the map uses classifications from GFSAD GCEP followed by CGLS-LC100-2016, as these were found to have the highest accuracy in the majority of AEZs in Africa. Using an independent random reference set to evaluate the combined map, Nabil et al. (2022) reported a precision score (user's accuracy) of $93.73\%$, recall score (producer's accuracy) of $61.93\%$, and overall accuracy of $91.64\%$.
We accessed the Nabil et al.~ map using the following GEE asset: \texttt{ee.Image("projects/sat-io/open-datasets/landcover/AF\_Cropland\_mask\_30m\_2016\_v3")}.

\paragraph{Majority Vote.} We created a majority vote ensemble to evaluate whether a combination of all of the maps compared in this study would achieve better performance than individual maps. If 6 or more of the 11 compared maps classified a pixel as crop, the pixel is classified as crop in the Majority Vote map if 5 or fewer classified a pixel as crop (i.e., 6 or more classified as non-crop) then the pixel is classified as non-crop. 

\subsection*{Reference datasets}
\subsubsection*{Sample design}
\begin{figure}[ht]
\centering
\includegraphics[width=\linewidth]{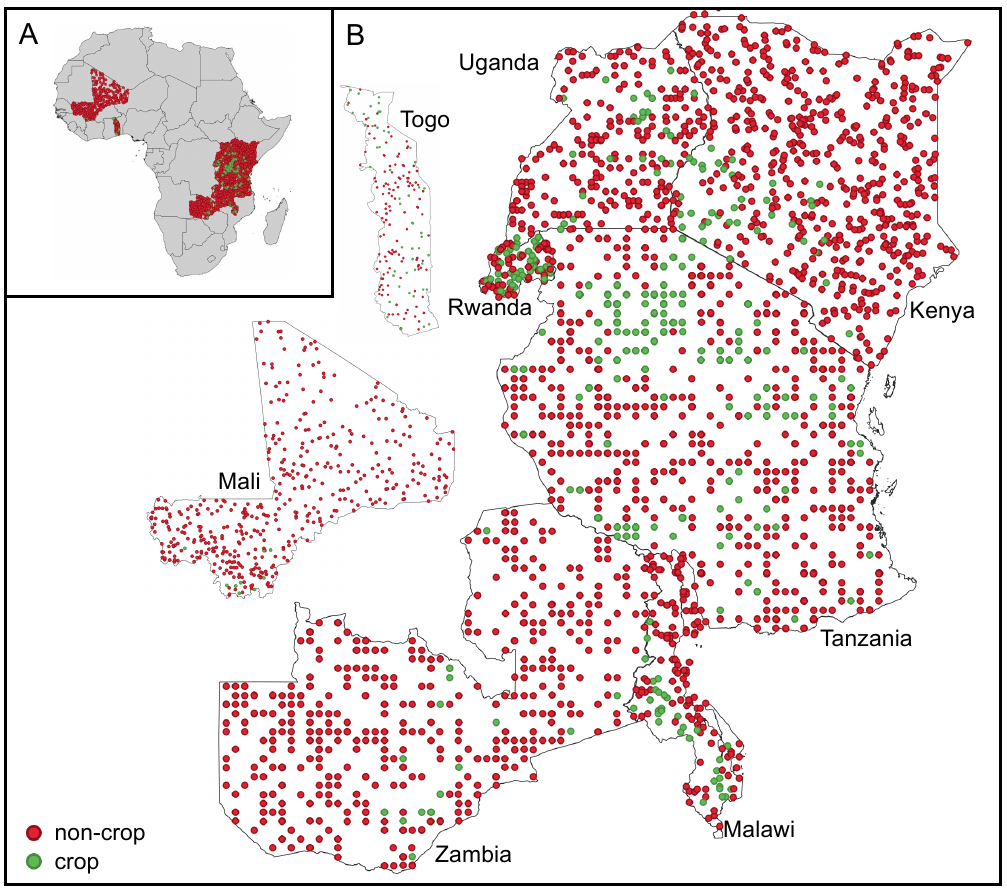}
\caption{Map showing the distribution of reference samples. Panel A shows the maps in the
context of the African continent while Panel B gives larger views of the maps (countries not to scale).}
\label{fig:reference-map}
\end{figure}

We created reference datasets for evaluating the accuracy of all maps in Togo, Kenya, Malawi, Mali, Rwanda, Tanzania, Uganda, and Zambia. As recommended by Stehman and Foody (2019),\cite{stehman2019key} we used a probability sampling design to ensure the representativeness of the sample and produce an unbiased estimate. For all countries except Mali, we sampled reference points by drawing a random uniform sample of point locations within each country's boundaries. Cropland constitutes a very small percentage of the total land area in Mali, thus a uniform random sample would result in a very small sample size for the cropland class. To try to increase the sample size for cropland in Mali, we sampled points using a stratified random sample with strata defined by four NDVI intervals using the mean annual NDVI from Sentinel-2: $(-1, 0.13], (0.13, 0.2], (0.2, 0.3], (0.3, 1]$. We assigned an equal number of points to each stratum. Table \ref{tab:evalsets} summarizes the characteristics of each dataset. Figure \ref{fig:reference-map} shows a map of the spatial distribution of the reference samples. 

\subsubsection*{Response design}
For each country sample, trained individuals manually analyzed high-resolution satellite images of PlanetScope (3 m/pixel resolution) monthly composites and other auxiliary sources (Sentinel-2, and sub-meter resolution images in Google Earth Pro) in the Collect Earth Online platform. For each point, annotators were instructed to inspect images from each month spanning the country's growing season and determine whether the point contained active cropland. We defined active cropland as points where patterns of sowing, growing, and/or harvesting in an agricultural field could be observed during the relevant agricultural season within a 12-month period. Interpreters were blind to the class predicted by any maps during labeling. At least two annotators labeled every point to maximize label confidence; the number of labelers that annotated each point in each dataset is indicated in the ``Num. labels per sample'' column in Table \ref{tab:evalsets}. We discarded points that did not have unanimous agreement between labelers to ensure high-confidence labels in the final reference dataset.

\begin{table}[ht]
\centering
\begin{tabular}{|l|cccc|c|c|}
\hline
{} & \multicolumn{4}{c|}{Number of samples} & {} & {} \\
Country & Total & Crop & Non-crop & Crop \% & Label validity period & Num. labels per sample\\
\hline\hline
Kenya & 573 & 36 & 537 & 6.3 & Feb 2019-Jan 2020 & 2 \\
Malawi & 154 & 31 & 123 & 25.2 & Sep 2020-Aug 2021 & 2 \\
Mali & 447 & 10 & 437 & 2.2 & Feb 2019-Jan 2020 & 2 \\ 
Rwanda & 107 & 43 & 64 & 40.2 & Jan 2019-Dec 2019 & 3 \\ 
Tanzania & 1201 & 431 & 770 & 35.9 & Jan 2019-Dec 2019 & 2 \\
Togo & 182 & 51 & 131 & 39.0 & Jan 2019-Dec 2019 & 4 \\
Uganda & 233 & 26 & 207 & 11.2 & Jan 2019-Dec 2019 & 2 \\
Zambia & 489 & 20 & 469 & 4.1 & Jan 2019-Dec 2019 & 2 \\
Total & 3386 & 740 & 2646 & 21.9 & -- & -- \\
\hline
\end{tabular}
\caption{\label{tab:evalsets}Reference sample datasets used for independent evaluation of cropland maps} 
\end{table}

\subsection*{Metrics and evaluation}

We used the following metrics to evaluate the accuracy of each map in this study using the reference datasets, which are commonly used for accuracy assessment in both machine learning and remote sensing: overall accuracy and F1 score, precision (also called user's accuracy or UA), and recall (also called producer's accuracy or PA). These metrics are reported in Table \ref{tab:metrics}. The recall for the positive (crop) class is also known as sensitivity. 
We used the equations from Stehman and Foody (2019)\cite{stehman2019key} and Olofsson et al. (2014)\cite{olofsson2014good} to compute estimates and standard errors for precision (UA), recall (PA), and overall accuracy. 
We computed the metrics from the population error matrix as recommended by Stehman and Foody (2019) to ensure the accuracy estimates are map relevant.\cite{stehman2019key} We describe in detail the procedure for computing the population error matrix and subsequent metrics. In the case of a binary map classification as in our study, the population error matrix is a $2\times2$ matrix in which rows represent the map classification and columns represent the reference classification expressed in terms of area proportion computed from the map. The population error matrix is computed from the sample error matrix, also called the confusion matrix. The confusion matrix $C$ is defined as follows in terms of the number of samples representing true negatives (TN), false negatives (FN), false positives (FP), and true positives (TP):
\begin{equation}
C=
\begin{bmatrix}
TN & FN \\
FP & TP 
\end{bmatrix}    
\end{equation}

The area matrix $A$ contains the total mapped area in each class expressed in terms of the number of pixels:
\begin{equation}
A=
\begin{bmatrix}
n_{neg} \\
n_{pos} 
\end{bmatrix}    
\end{equation}

The total number of pixels in the map is $n=n_{neg}+n_{pos}$. The weight matrix $W$ gives the proportion of area mapped as each class, computed by dividing each element in the area matrix by the total mapped area:
\begin{equation}
    W = A / n
\end{equation}

The population error matrix $E$ can then be computed as follows, where $C_{\bullet j}$ is the sum of each column in the confusion matrix $C$:
\begin{equation}
    E = \frac{W*C}{C_{\bullet j}}
\end{equation}

Overall accuracy, precision, and recall can be computed using the elements $p_{ij}$ of the population error matrix $E$. Overall accuracy is computed by summing the diagonal elements of the population error matrix:  
\begin{equation}
    OA = \sum_{i=1}^c{p_{ii}} 
\end{equation}
where $c$ is the number of classes. The precision (user's accuracy) for each class can be computed as:
\begin{equation}
    P_i = \frac{p_{ii}}{p_{i\bullet}}
\end{equation}
where $p_{i\bullet}$ is the sum of each column in $E$. The recall (producer's accuracy) for each class can be computed as:
\begin{equation}
    R_j = \frac{p_{jj}}{p_{\bullet j}}
\end{equation}
where $p_{\bullet j}$ is the sum of each row in $E$.

The variance estimators for each metric are given by Equations 16-18 in Stehman and Foody (2019)\cite{stehman2019key}, which we re-write using the terminology of the previous equations as follows:
\begin{align}
    V(OA) &= \sum_{i=1}^c {\frac{W_i^2 P_i ( 1 - P_i)}{C_{i\bullet} - 1}} \\
    V(P_i) &= \frac{W_i^2 P_i ( 1 - P_i)}{C_{i\bullet} - 1} \\
    V(R_j) &= \frac{1}{\hat{N}_{\bullet j}^2} 
    [ \frac{A_{j}^2 (1-R_j)^2 P_j(1-P_j)}{C_{j\bullet}-1} + 
    R_j^2 \sum_{i\neq j}^c A_{i\bullet}^2 \frac{C_{ij}}{C_{i\bullet}} (1-\frac{C_{ij}}{C_{i\bullet}}) / (C_{i\bullet} - 1)]
\end{align}
where $\hat{N}_{\bullet j}$ is the estimated total number of pixels of reference class $j$.
In our results, we reported the standard deviation $\sigma$ calculated from each variance estimate $v$: $\sigma = \sqrt{v}$.
We computed the F1 score using the estimates of precision and recall and computed the standard error using error propagation\cite{uncertainty} as follows:
\begin{align}
F_1 &= \frac{2PR}{P+R} \text{, which can be simplified to } \\
F_1 &= \frac{X}{Y} \text{, where } X = 2PR \text{ and } Y = P+R
\end{align}
Since the uncertainty of X and Y are not independent, they do not add in quadrature. Instead, their relative errors are added:
\begin{align}
    \frac{\Delta F_1}{F_1}  &= \frac{\Delta X}{X} + \frac{\Delta Y}{Y}\text{, where } \Delta F_1, \Delta X, \Delta Y \text{are the errors of } F_1, X, Y 
\end{align}
Since $F_1=X/Y$, we multiply both sides by $X/Y$ to get:
\begin{align}    
    \Delta F_1 &= \frac{\Delta X}{Y} + \frac{X\Delta Y}{YY} 
\end{align}
The error in $X$ and $Y$ are:
\begin{align} 
\Delta X &= 2(R\Delta P + P\Delta R)\\
\Delta Y &= \Delta P + \Delta R
\end{align}
We substitute Eqns. 8 and 9 into Eqn. 7 to get the final expression for the error of F1:
\begin{align} 
         \Delta F_1 &= \frac{2(R\Delta P + P\Delta R)}{P+R} + \frac{2PR(\Delta P + \Delta R)}{(P+R)(P+R)}
\end{align}

\section*{Data availability}
The datasets used in this analysis are available from Zenodo.\cite{kerner_2024_10694610} The Zenodo repository includes the following specific datasets:
\begin{enumerate}
    \item Labeled reference datasets for each country used for evaluation and summarized in Table \ref{tab:evalsets}, in shapefile format
    \item CSV file of the evaluation metrics computed for each map using the reference dataset from each country
    \item Consensus maps for each country indicating in each pixel the number of maps that predict cropland (used to create Figure \ref{fig:agreement}), in GeoTIFF format with 10 m/pixel resolution
\end{enumerate}

The reference datasets are also available as public feature collection assets via Google Earth Engine (note that you must be signed in with a Google Earth Engine account to view/access these links): 
\begin{enumerate}
    \item \url{https://code.earthengine.google.com/?asset=projects/bsos-geog-harvest1/assets/harvest-reference-datasets/kenya-2019-reference-points}
    \item \url{https://code.earthengine.google.com/?asset=projects/bsos-geog-harvest1/assets/harvest-reference-datasets/malawi-2020-reference-points}
    \item \url{https://code.earthengine.google.com/?asset=projects/bsos-geog-harvest1/assets/harvest-reference-datasets/mali-2019-reference-points}
    \item \url{https://code.earthengine.google.com/?asset=projects/bsos-geog-harvest1/assets/harvest-reference-datasets/rwanda-2019-reference-points}
    \item \url{https://code.earthengine.google.com/?asset=projects/bsos-geog-harvest1/assets/harvest-reference-datasets/tanzania-2019-reference-points}
    \item \url{https://code.earthengine.google.com/?asset=projects/bsos-geog-harvest1/assets/harvest-reference-datasets/togo-2019-reference-points}
    \item \url{https://code.earthengine.google.com/?asset=projects/bsos-geog-harvest1/assets/harvest-reference-datasets/uganda-2019-reference-points}
    \item \url{https://code.earthengine.google.com/?asset=projects/bsos-geog-harvest1/assets/harvest-reference-datasets/zambia-2019-reference-points}
\end{enumerate}

The maps and other datasets used in this study can be visualized and compared in a Google Earth Engine app at \url{https://hkerner_umd.users.earthengine.app/view/intercomparison-of-public-crop-maps-in-sub-saharan-africa}.

\section*{Code availability}
The code used for processing and evaluating each of the maps in this study is publicly accessible at \url{https://github.com/nasaharvest/crop-mask/blob/master/src/compare_covermaps.py}.

\bibliography{intercomparison}

\section*{Acknowledgements}

We are grateful to Dr. Samapriya Roy for adding many of the map products used in this study to the \texttt{awesome-gee-datasets} catalog and to Dr. Jacob Adler for assistance with data visualization and error estimation. This work was supported by the NASA Harvest Consortium on Food Security and Agriculture (Award \#80NSSC18M0039). We are also grateful to the many individuals who contributed to labeling the reference datasets used in this study.

\section*{Author contributions statement}

H.K. conceived the study and experiment(s), H.K. and A.Y. conducted the experiments, H.K. analyzed the results. The manuscript was written by H.K. and C.N. All authors reviewed the manuscript and provided input on results and experiments. 

\section*{Competing interests}

The authors declare no competing interests.

\end{document}


\section*{Supplementary File 1}

\begin{figure}[ht]
\centering
\includegraphics[width=\linewidth]{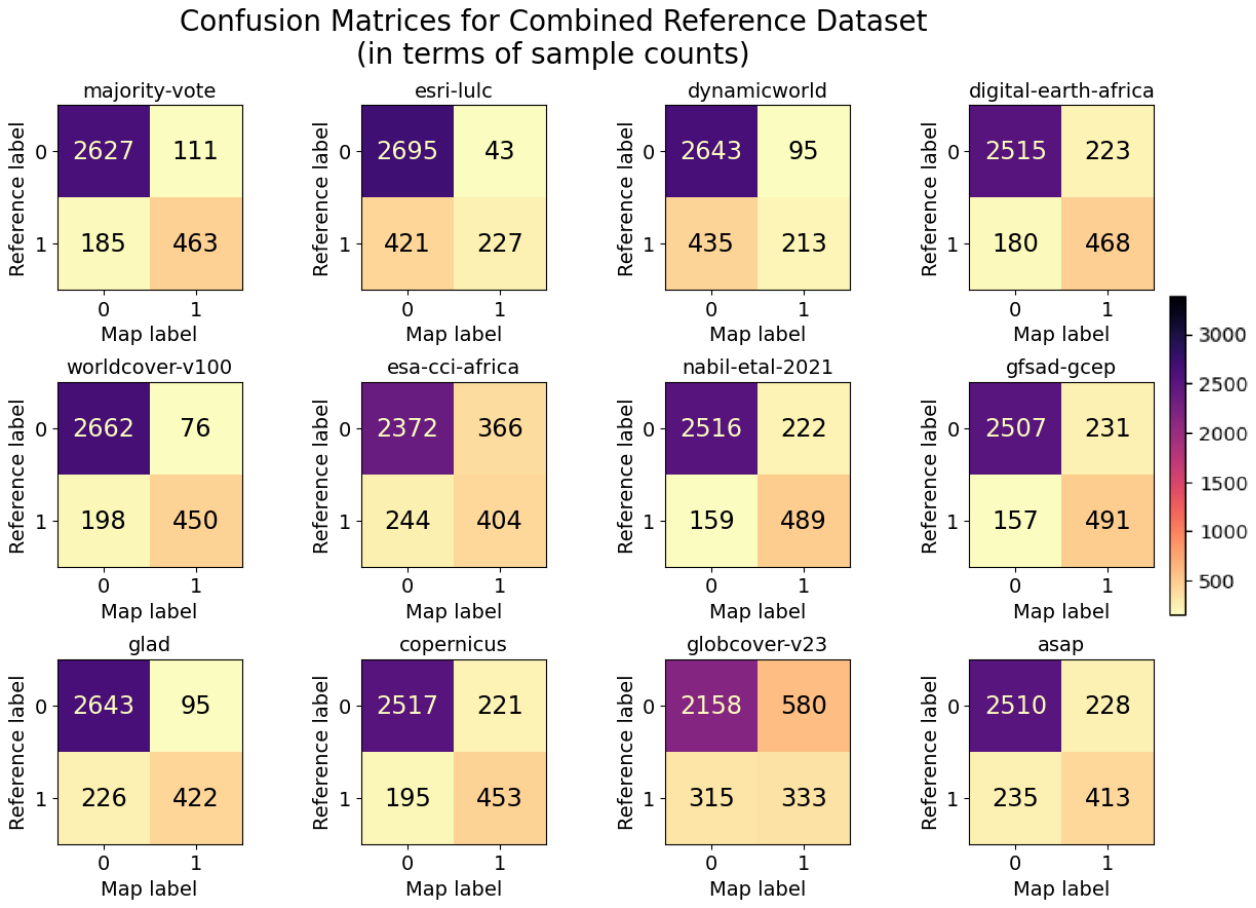}
\caption{Unnormalized confusion matrix, expressed in terms of sample counts, for each map for the combined reference datasets.}
\label{fig:cm-combined}
\end{figure}

\begin{figure}[ht]
\centering
\includegraphics[width=\linewidth]{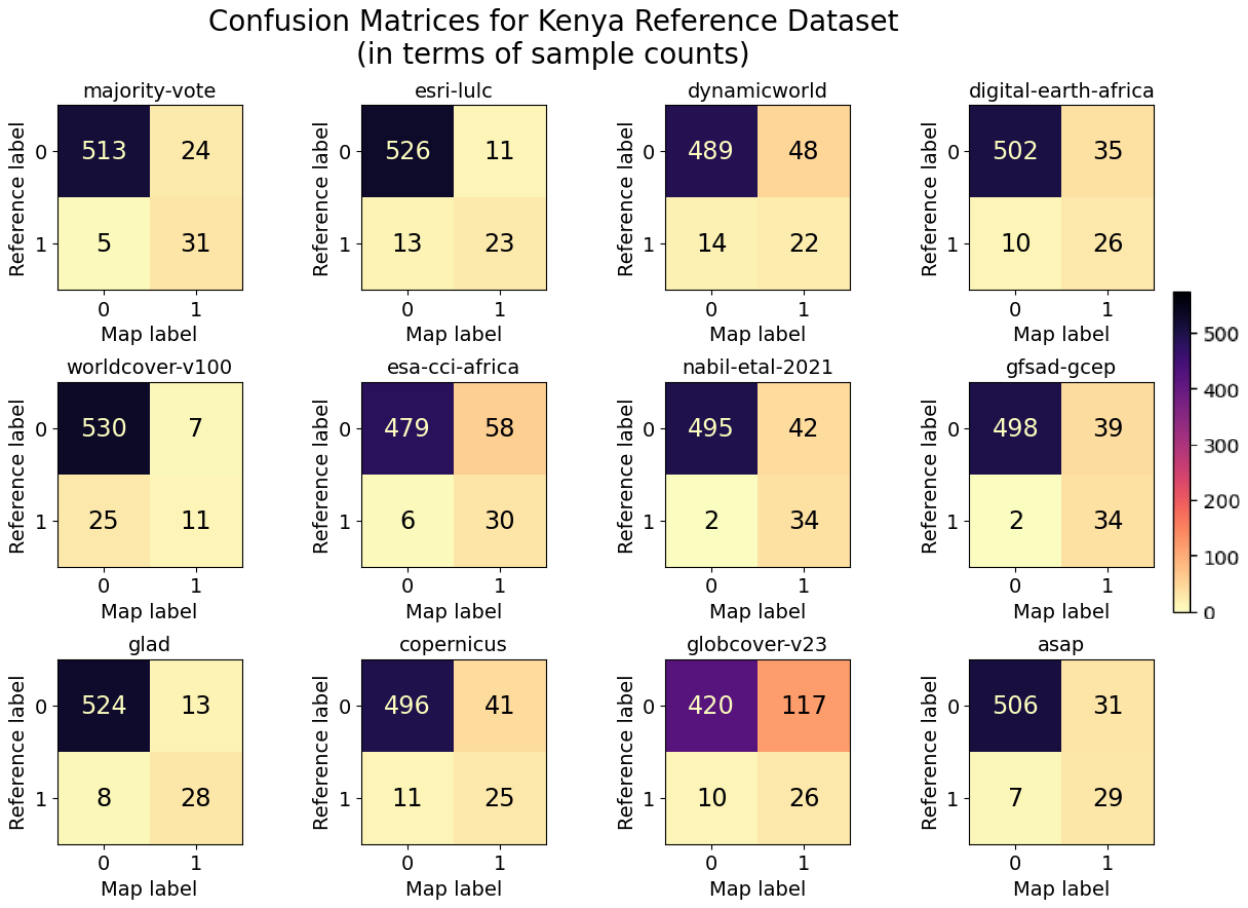}
\caption{Unnormalized confusion matrix, expressed in terms of sample counts, for the Kenya reference dataset.}
\label{fig:cm-kenya}
\end{figure}

\begin{figure}[ht]
\centering
\includegraphics[width=\linewidth]{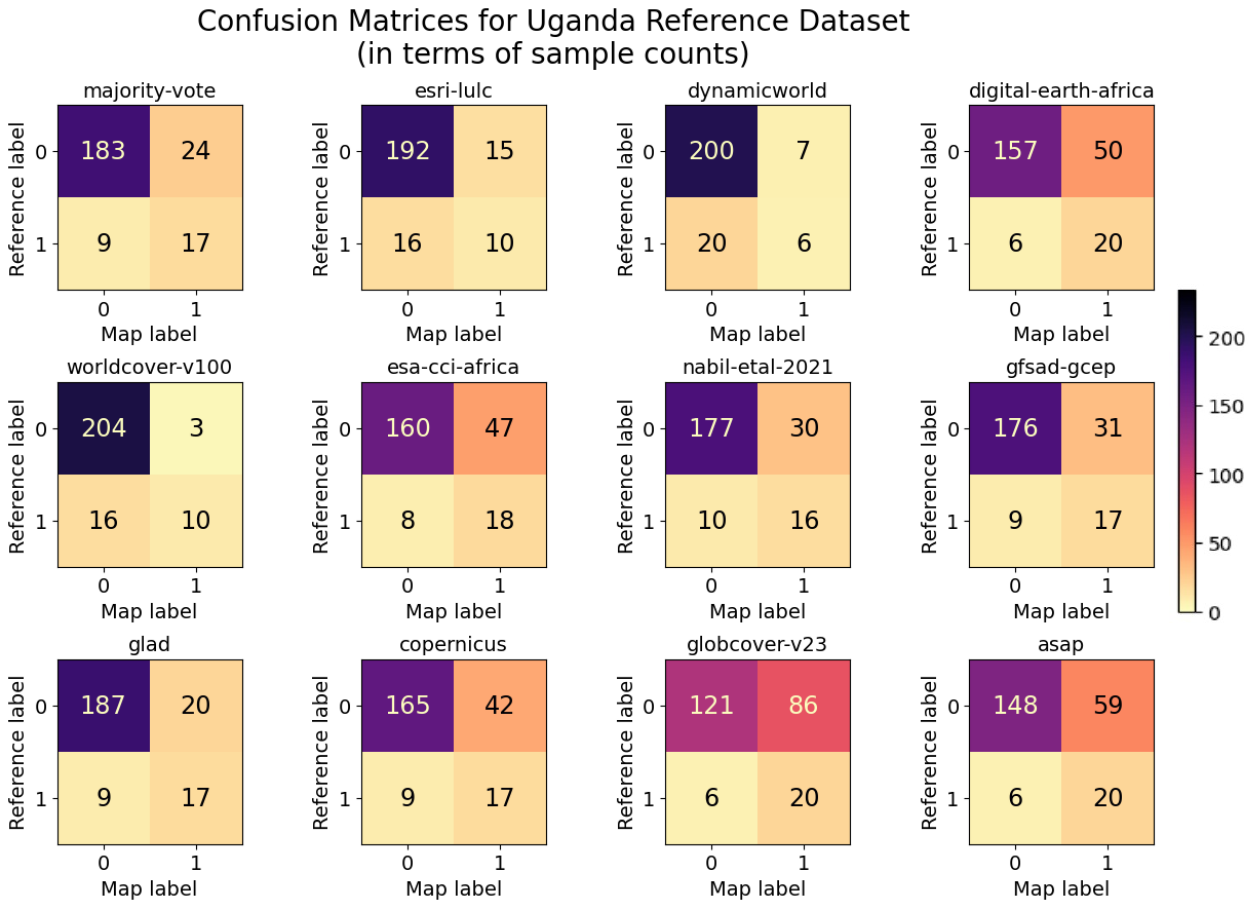}
\caption{Unnormalized confusion matrix, expressed in terms of sample counts, for the Uganda reference dataset.}
\label{fig:cm-uganda}
\end{figure}

\begin{figure}[ht]
\centering
\includegraphics[width=\linewidth]{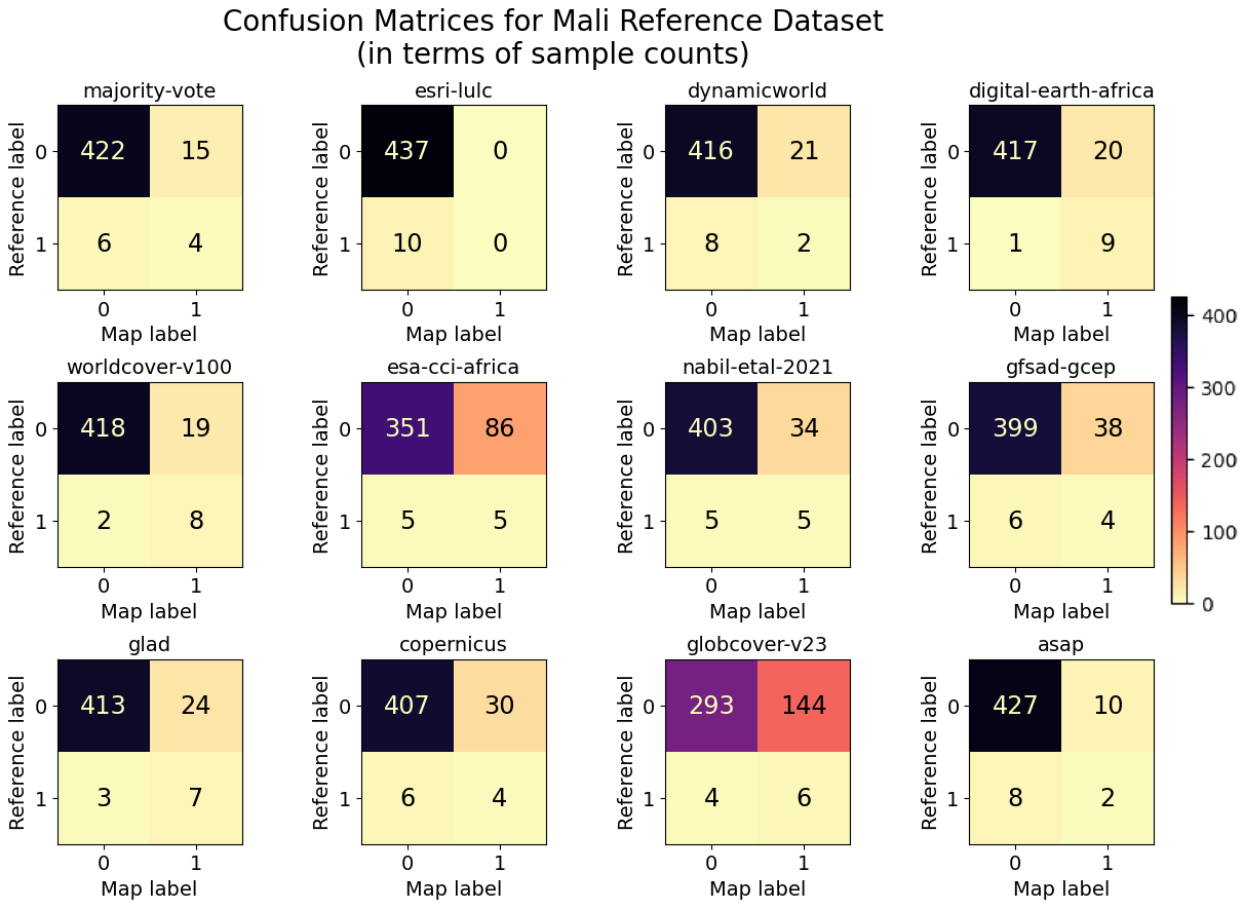}
\caption{Unnormalized confusion matrix, expressed in terms of sample counts, for the Mali reference dataset.}
\label{fig:cm-mali}
\end{figure}

\begin{figure}[ht]
\centering
\includegraphics[width=\linewidth]{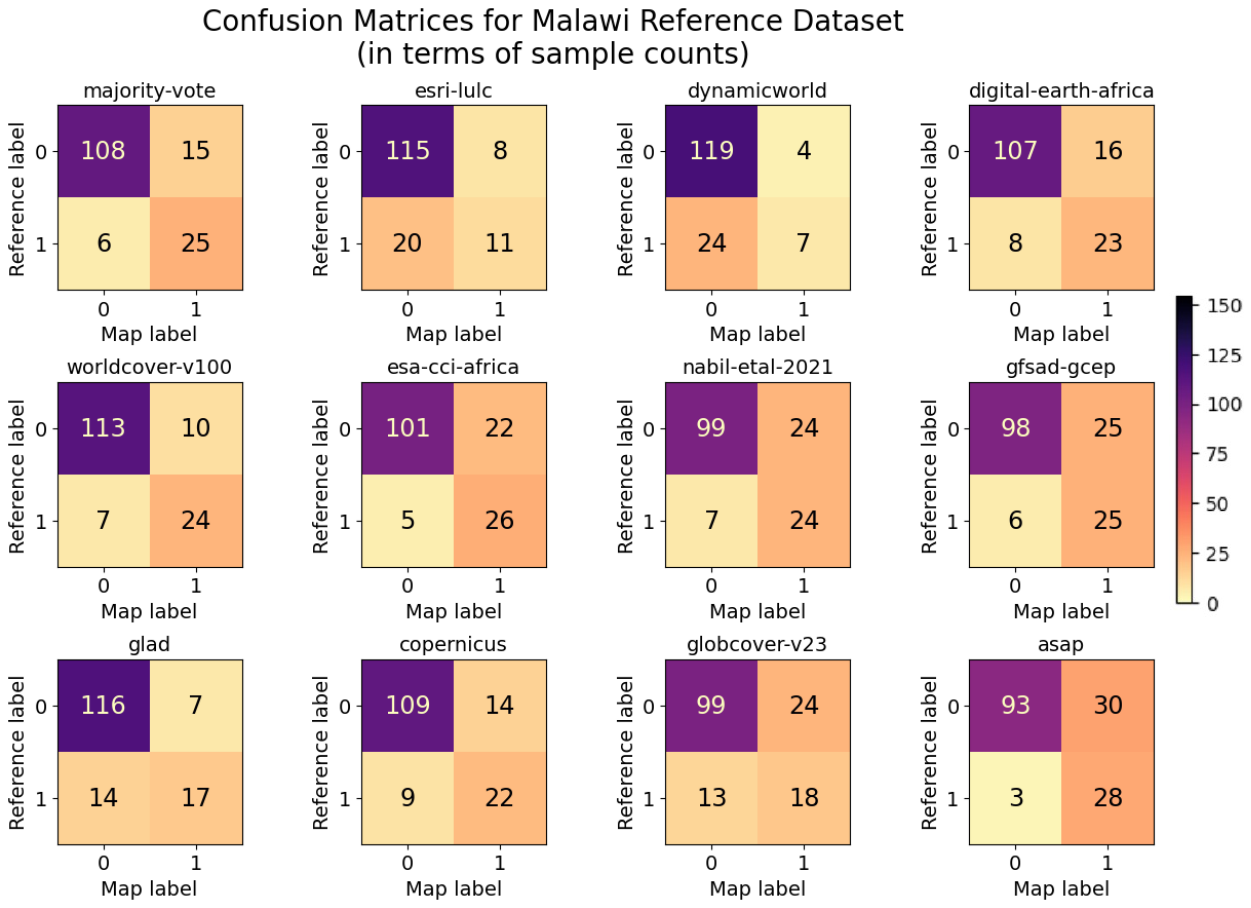}
\caption{Unnormalized confusion matrix, expressed in terms of sample counts, for the Malawi reference dataset.}
\label{fig:cm-malawi}
\end{figure}

\begin{figure}[ht]
\centering
\includegraphics[width=\linewidth]{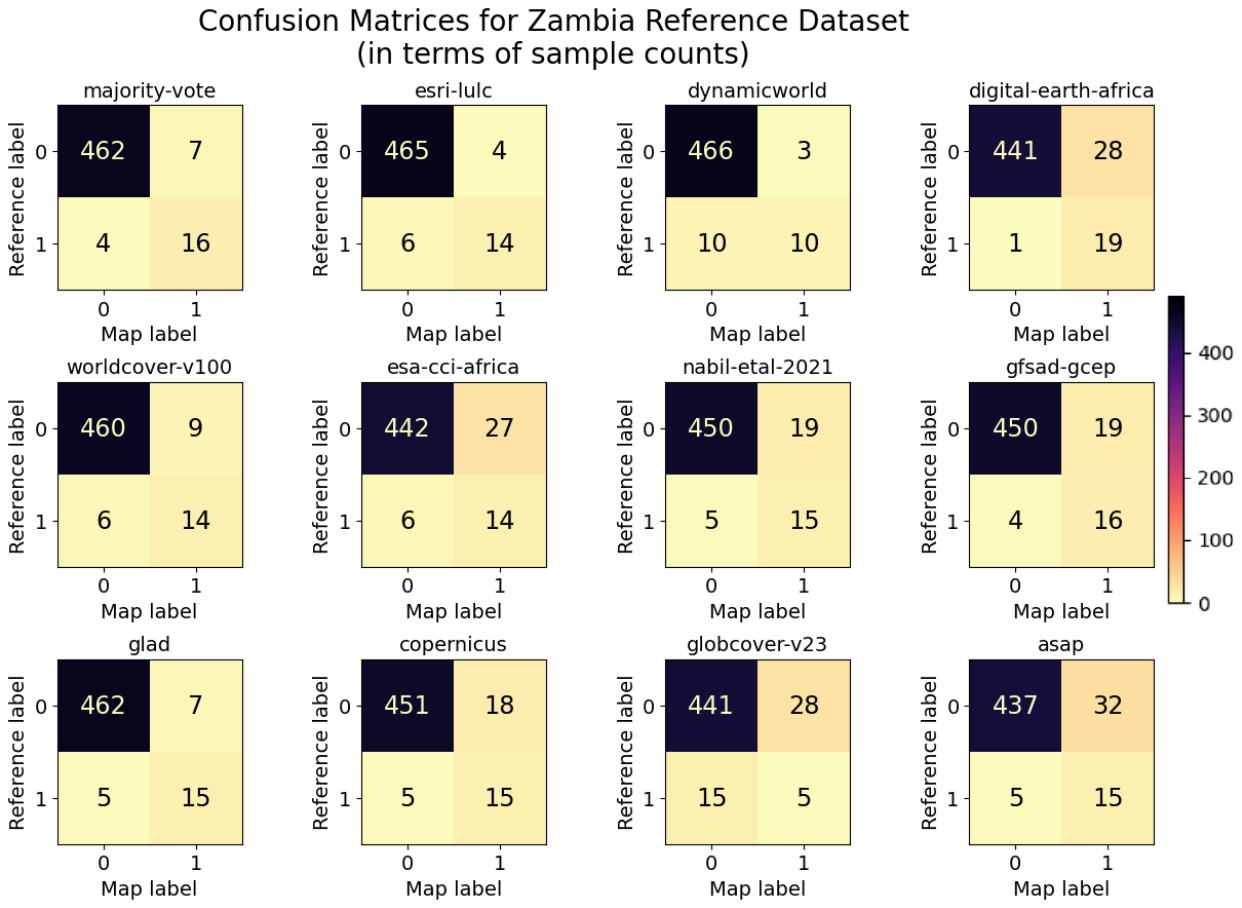}
\caption{Unnormalized confusion matrix, expressed in terms of sample counts, for the Zambia reference dataset.}
\label{fig:cm-zambia}
\end{figure}

\begin{figure}[ht]
\centering
\includegraphics[width=\linewidth]{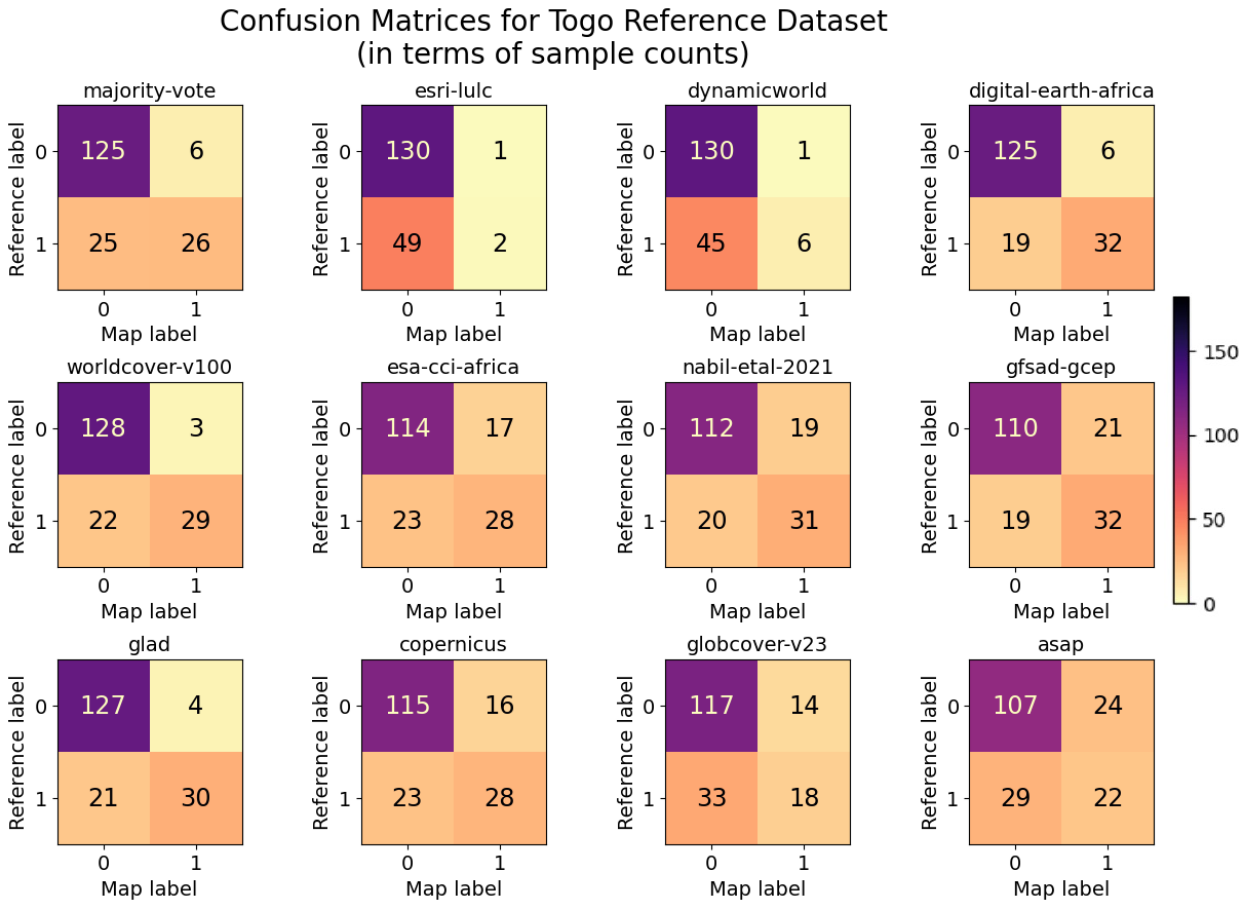}
\caption{Unnormalized confusion matrix, expressed in terms of sample counts, for the Togo reference dataset.}
\label{fig:cm-togo}
\end{figure}

\begin{figure}[ht]
\centering
\includegraphics[width=\linewidth]{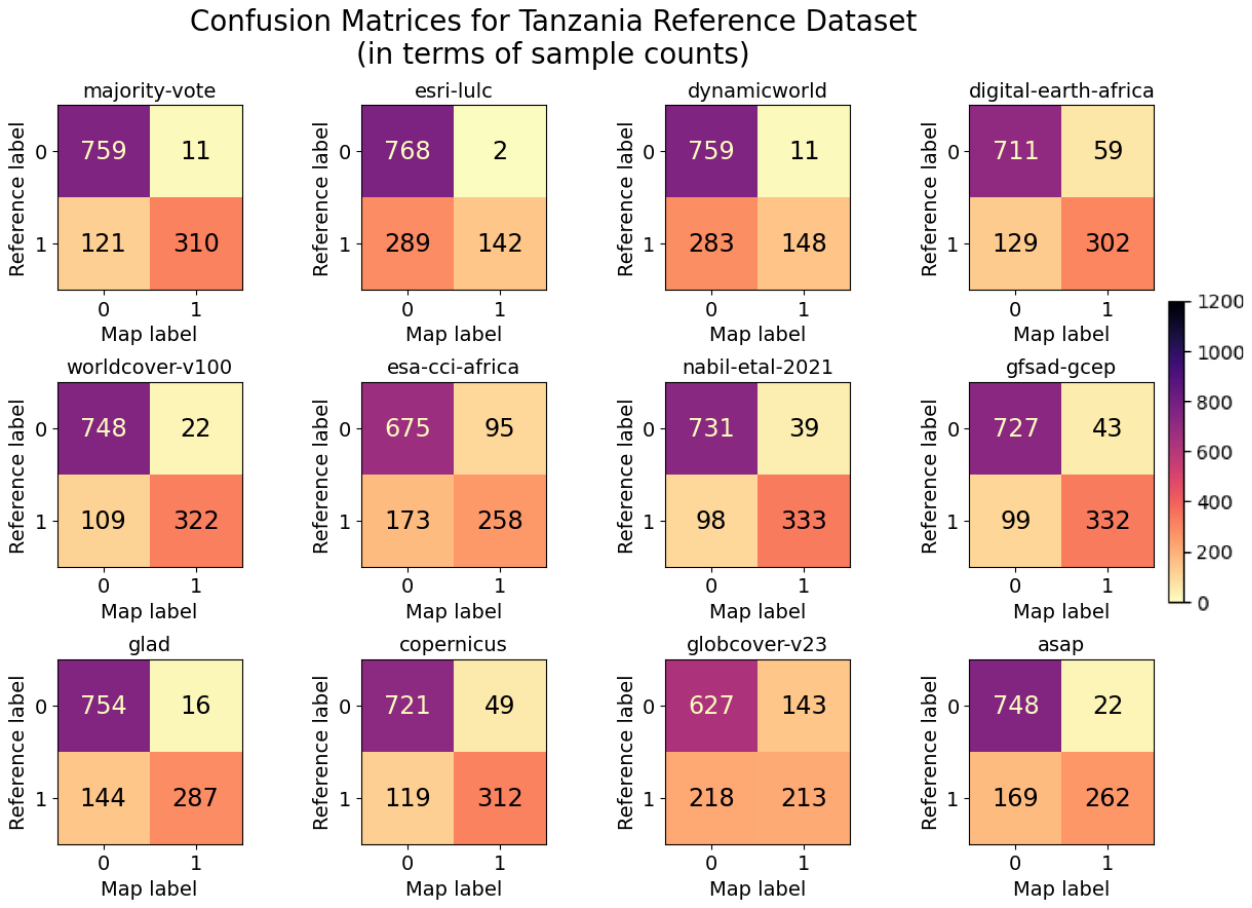}
\caption{Unnormalized confusion matrix, expressed in terms of sample counts, for the Tanzania reference dataset.}
\label{fig:cm-tanzania}
\end{figure}

\begin{figure}[ht]
\centering
\includegraphics[width=\linewidth]{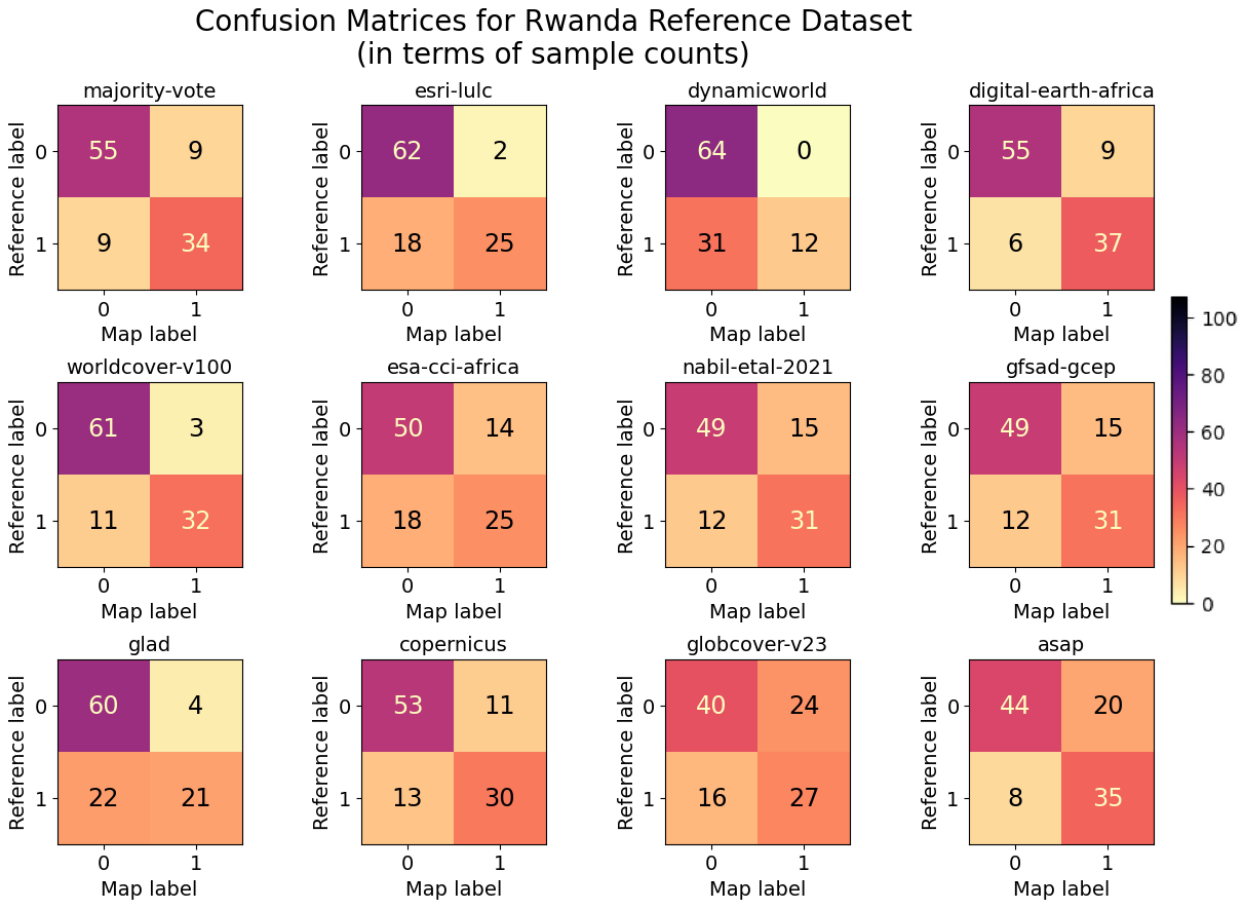}
\caption{Unnormalized confusion matrix, expressed in terms of sample counts, for the Rwanda reference dataset.}
\label{fig:cm-rwanda}
\end{figure}

\begin{figure}[ht]
\centering
\includegraphics[width=\linewidth]{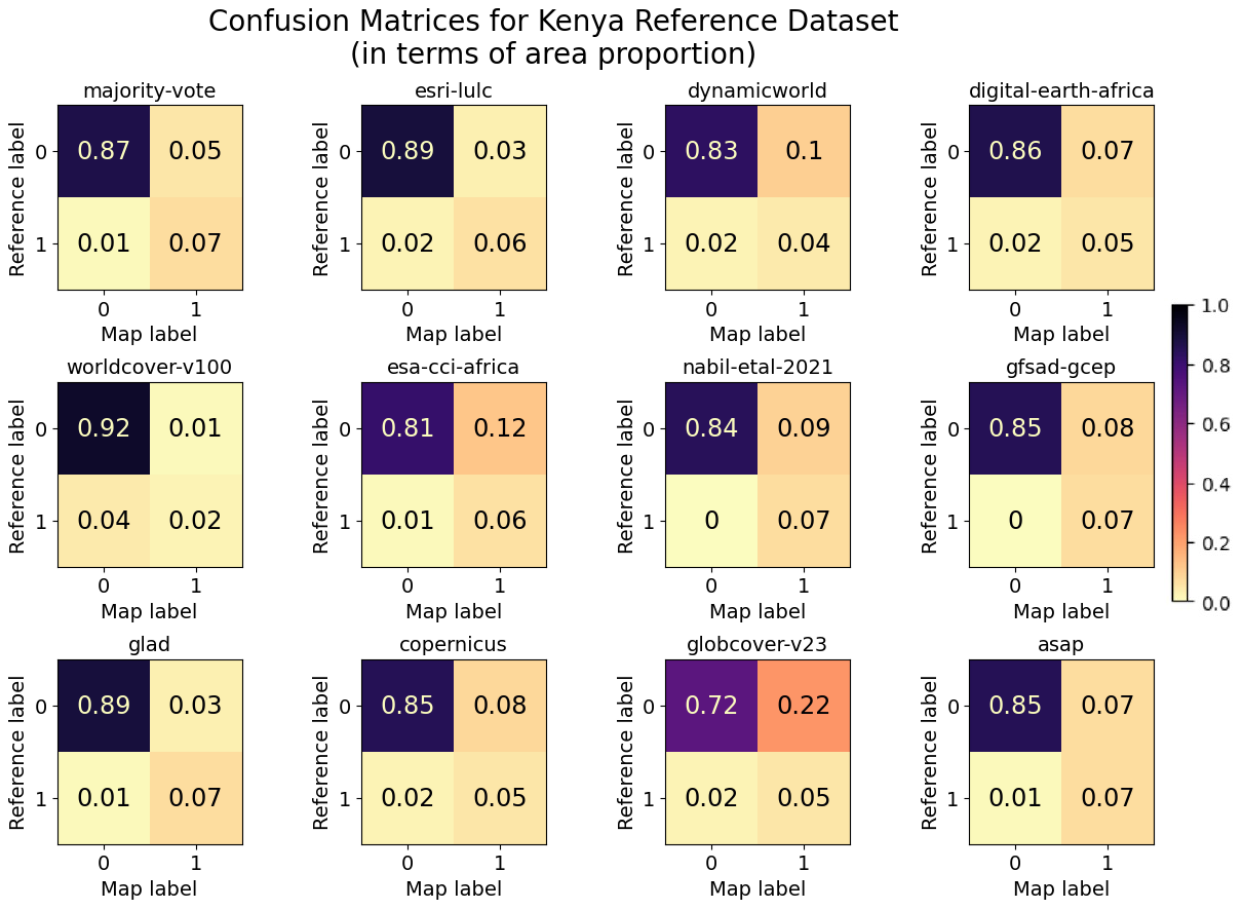}
\caption{Population confusion matrix, expressed in terms of map area proportion, for the Kenya reference dataset.}
\label{fig:cm-kenya}
\end{figure}

\begin{figure}[ht]
\centering
\includegraphics[width=\linewidth]{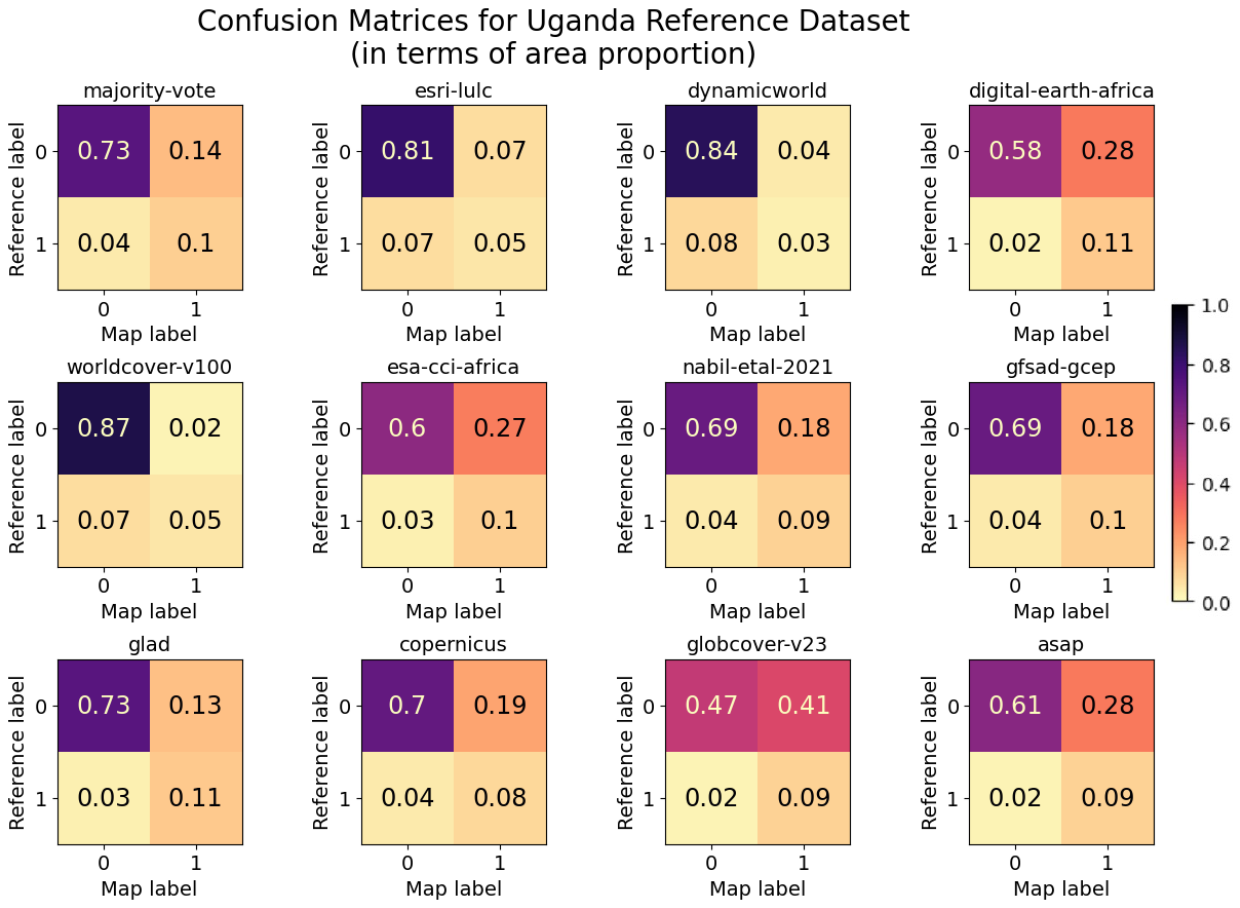}
\caption{Population confusion matrix, expressed in terms of map area proportion, for the Uganda reference dataset.}
\label{fig:cm-uganda}
\end{figure}

\begin{figure}[ht]
\centering
\includegraphics[width=\linewidth]{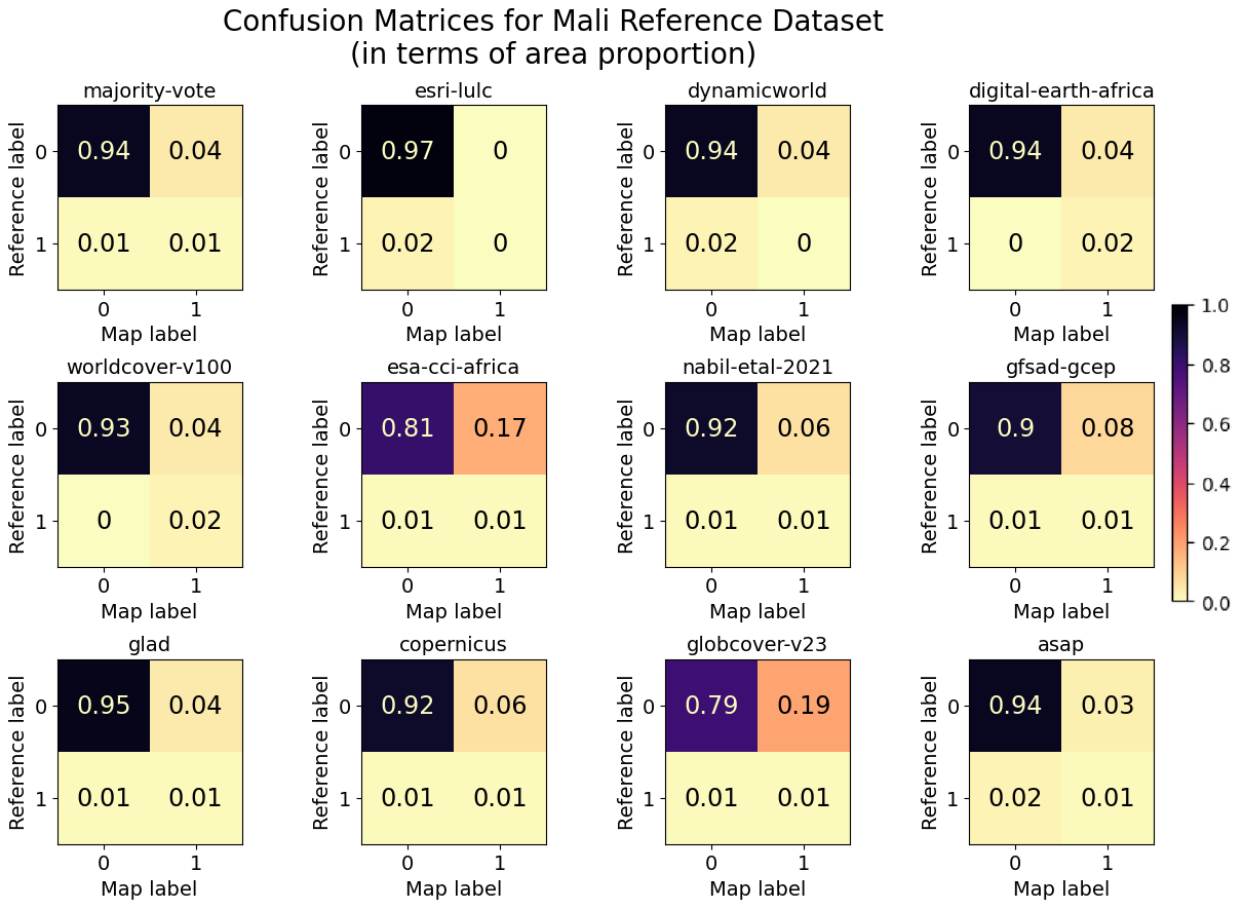}
\caption{Population confusion matrix, expressed in terms of map area proportion, for the Mali reference dataset.}
\label{fig:cm-mali}
\end{figure}

\begin{figure}[ht]
\centering
\includegraphics[width=\linewidth]{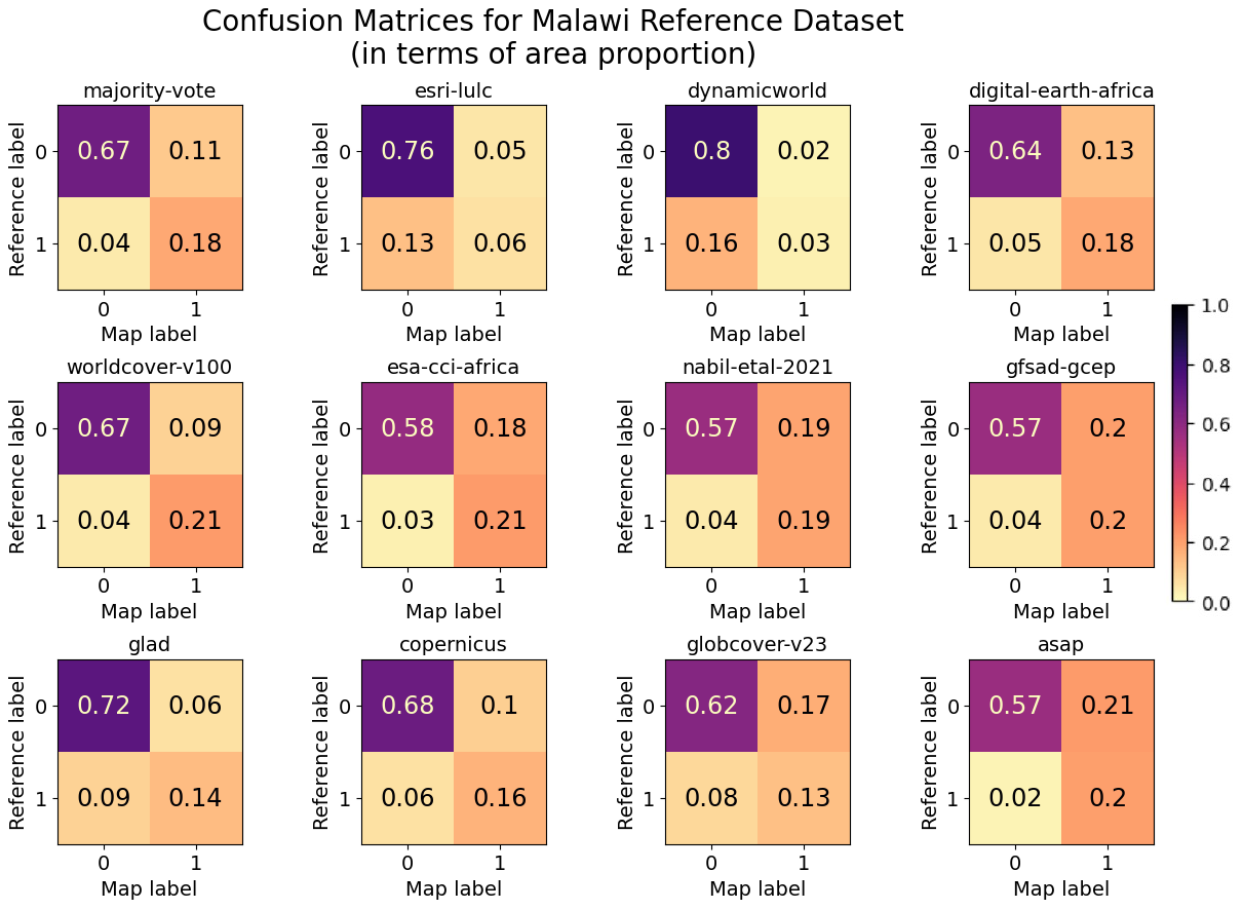}
\caption{Population confusion matrix, expressed in terms of map area proportion, for the Malawi reference dataset.}
\label{fig:cm-malawi}
\end{figure}

\begin{figure}[ht]
\centering
\includegraphics[width=\linewidth]{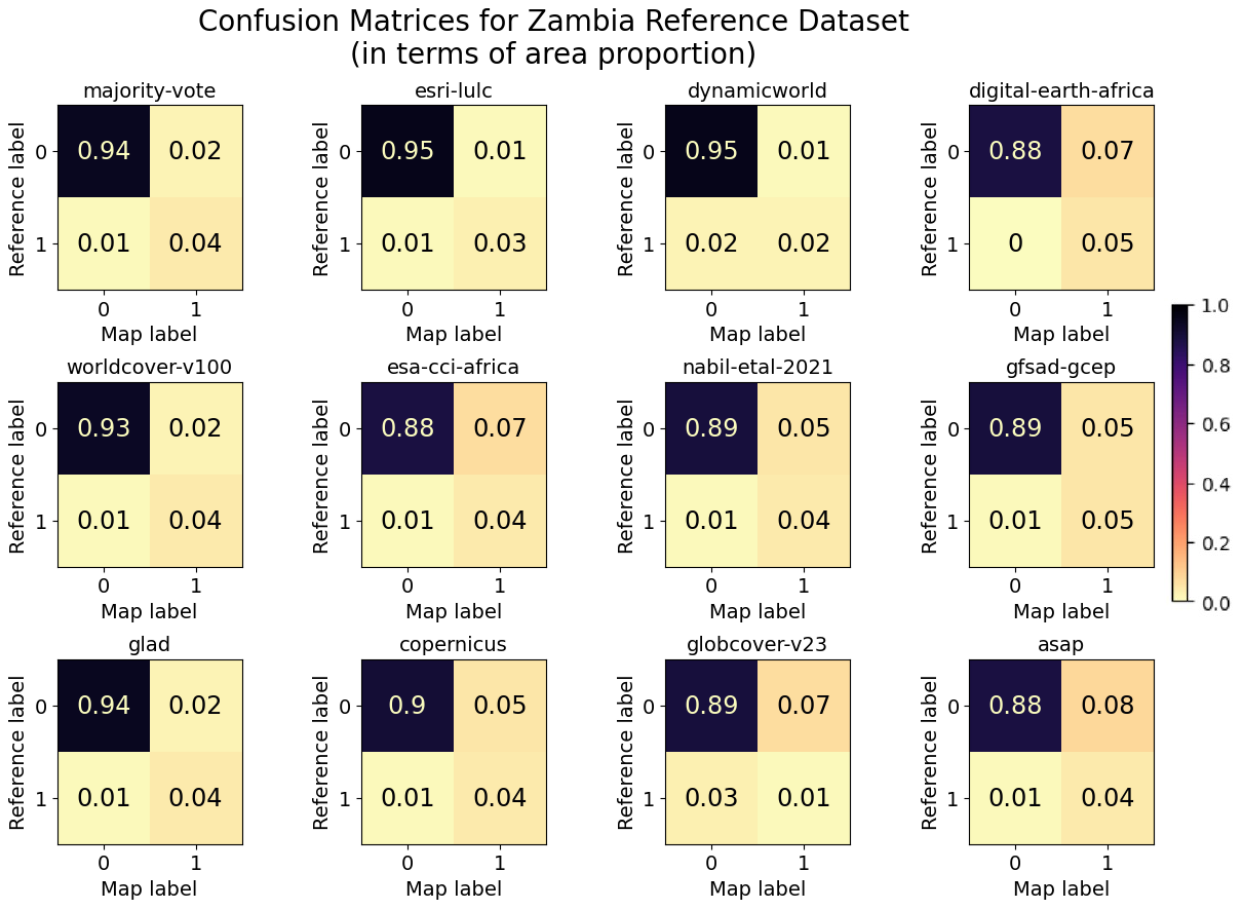}
\caption{Population confusion matrix, expressed in terms of map area proportion, for the Zambia reference dataset.}
\label{fig:cm-zambia}
\end{figure}

\begin{figure}[ht]
\centering
\includegraphics[width=\linewidth]{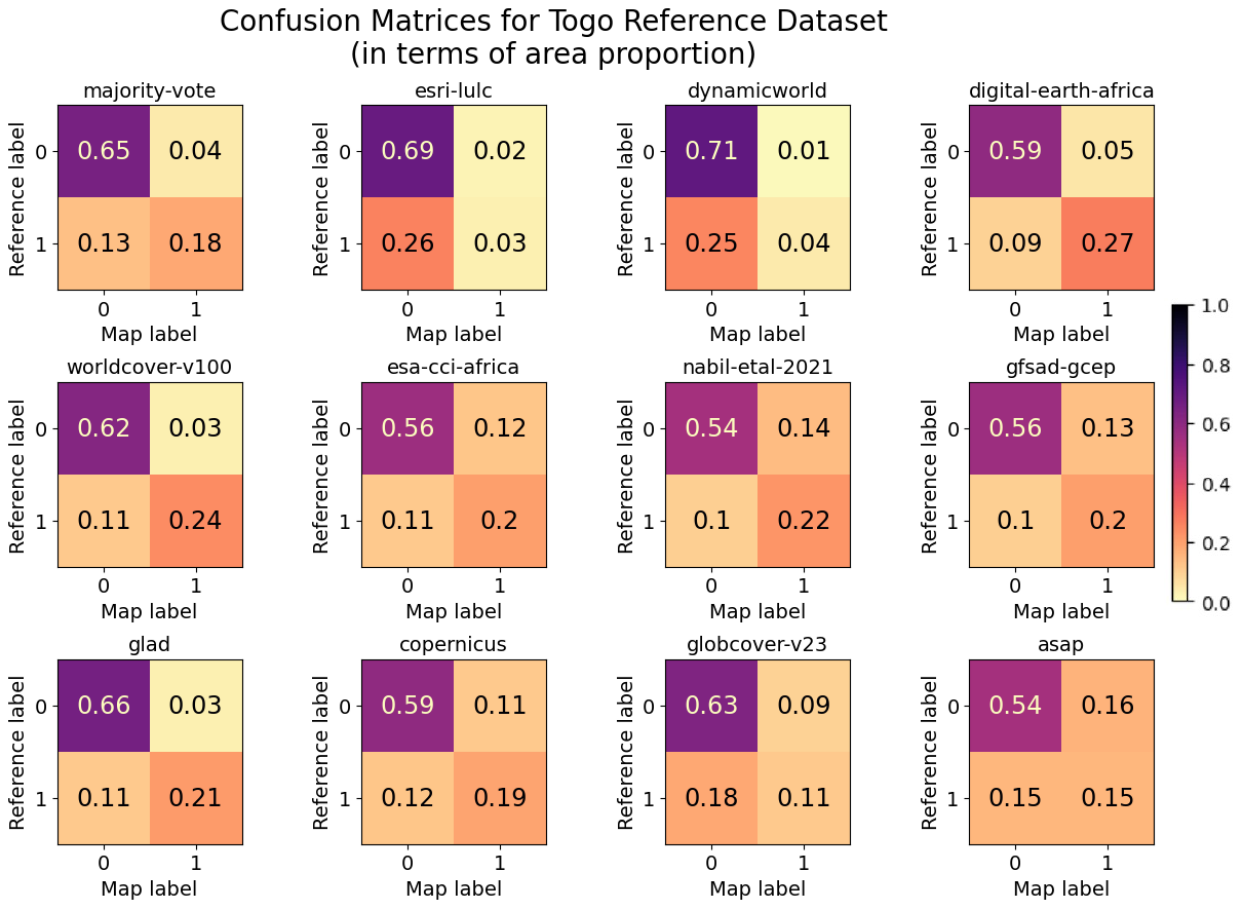}
\caption{Population confusion matrix, expressed in terms of map area proportion, for the Togo reference dataset.}
\label{fig:cm-togo}
\end{figure}

\begin{figure}[ht]
\centering
\includegraphics[width=\linewidth]{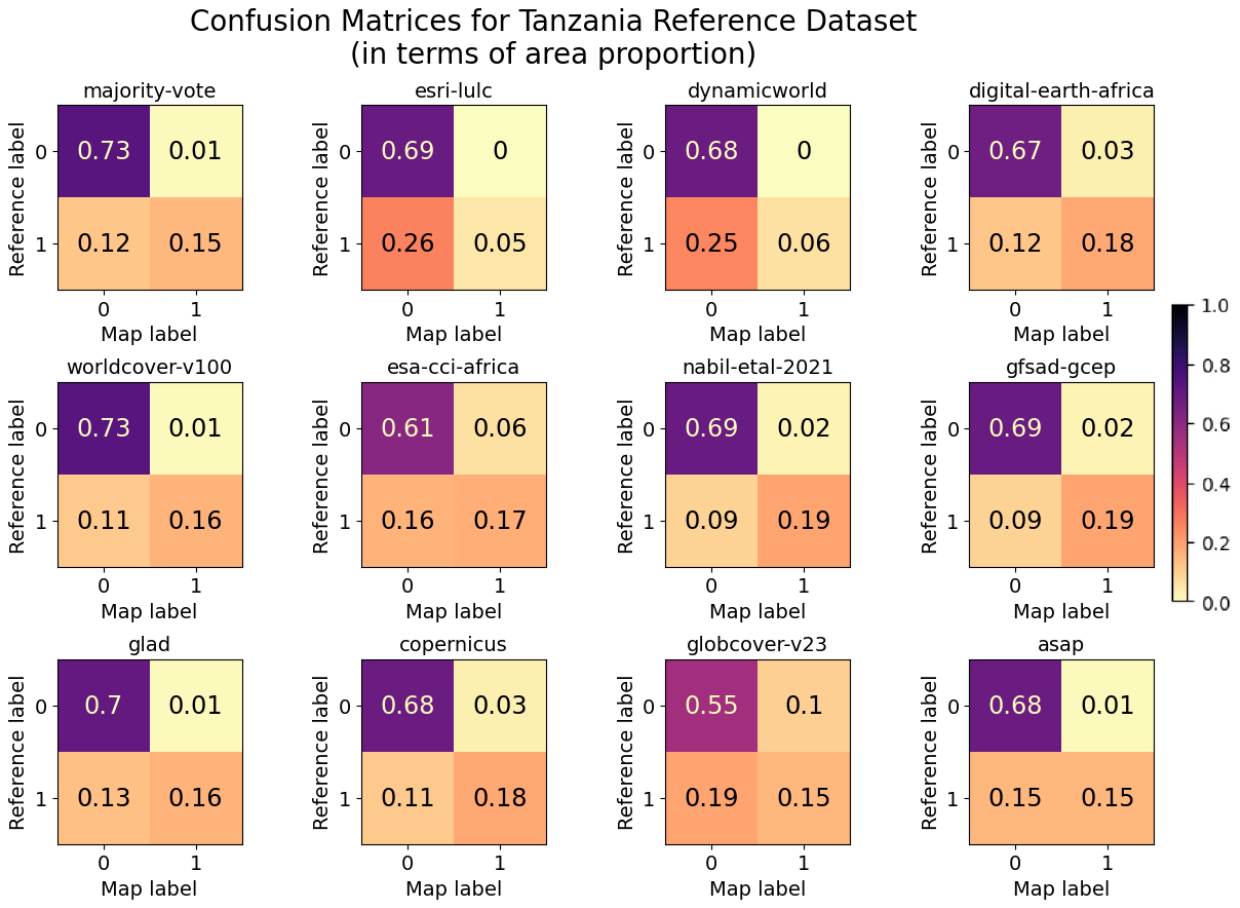}
\caption{Population confusion matrix, expressed in terms of map area proportion, for the Tanzania reference dataset.}
\label{fig:cm-tanzania}
\end{figure}

\begin{figure}[ht]
\centering
\includegraphics[width=\linewidth]{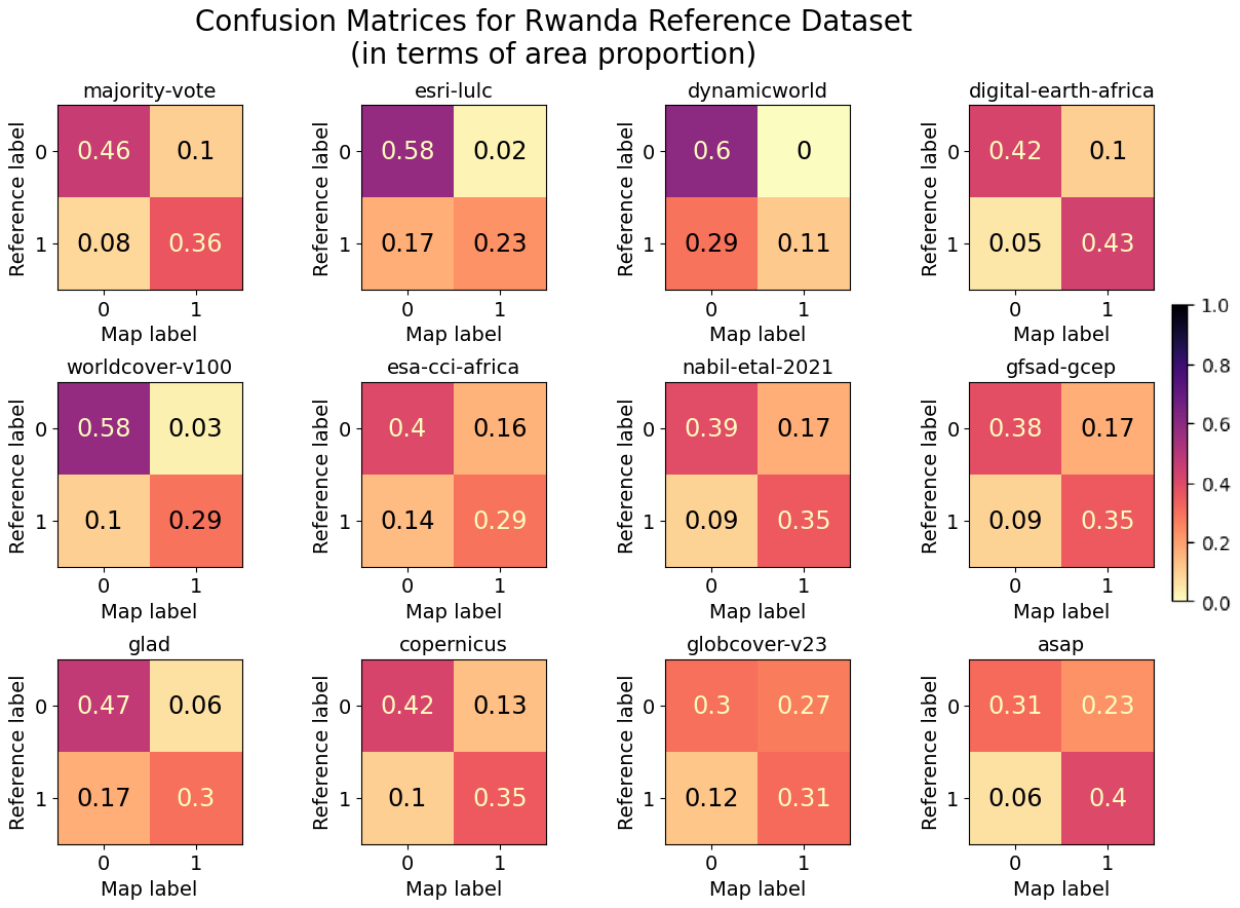}
\caption{Population confusion matrix, expressed in terms of map area proportion, for the Rwanda reference dataset.}
\label{fig:cm-rwanda}
\end{figure}